\documentclass[]{amazonpreprint}
\usepackage[utf8]{inputenc}
\usepackage[T1]{fontenc} 
\usepackage{hyperref}   
\usepackage{url}        
\usepackage{booktabs}       
\usepackage{amsfonts}       
\usepackage{nicefrac}      
\usepackage{microtype}     
\usepackage{xcolor}  
\definecolor{md-bg}{HTML}{F5F5F5} 
\usepackage{tikz}
\usepackage{algorithm}
\usepackage{algorithmic} 
\usepackage{listings}
\usepackage{caption}
\usepackage{amsthm}
\usepackage{amsmath}
\usepackage[inline]{enumitem}
\usepackage{xspace} 
\usepackage{cleveref}
\usepackage{subscript}
\usepackage{fvextra}
\usepackage{framed}
\usepackage{color}
\usepackage{graphicx} 
\usepackage{wrapfig} 
\usepackage{subcaption}
\usepackage{mathtools}
\usepackage{adjustbox}
\usepackage{siunitx}
\usepackage{twemojis}
\usepackage{tabularx}
\usepackage{fontawesome5}
\usepackage{amssymb}

\usepackage[labelsep=colon]{caption}
\usepackage[normalem]{ulem}  
\definecolor{shadecolor}{rgb}{0.95, 0.95, 0.95}

\definecolor{lightgray}{gray}{0.95}
\definecolor{darkblue}{rgb}{0.1,0.2,0.6}
\definecolor{codegray}{rgb}{0.4,0.4,0.4}

\usetikzlibrary{shapes.geometric, arrows.meta, positioning, fit, calc, backgrounds, shadows}

\definecolor{primaryBlue}{RGB}{0, 105, 148}
\definecolor{workerGreen}{RGB}{46, 139, 87}
\definecolor{evalOrange}{RGB}{230, 126, 34}
\definecolor{outputGold}{RGB}{241, 196, 15}
\definecolor{darkGray}{RGB}{50, 50, 50}

\usepackage{natbib}

\usepackage{xcolor}
\usepackage{tikz}
\usetikzlibrary{positioning, arrows.meta, calc}

\definecolor{lvlone}{RGB}{31, 119, 180}
\definecolor{lvltwo}{RGB}{214, 97, 43}
\definecolor{lvlthree}{RGB}{117, 82, 160}



\lstdefinestyle{pddlstyle}{
  numbers=left,
  numbersep=-6pt,
  backgroundcolor=\color{lightgray},
  basicstyle=\ttfamily\scriptsize,
  keywordstyle=\color{darkblue}\bfseries,
  commentstyle=\color{codegray}\itshape,
  showstringspaces=false,
  frame=single,
  breaklines=true,
  language=PDDL
}

\lstdefinestyle{nl}{
  numbers=left,
  numbersep=-6pt,
  backgroundcolor=\color{lightgray},
  basicstyle=\ttfamily\scriptsize,
  keywordstyle=\color{darkblue}\bfseries,
  commentstyle=\color{codegray}\itshape,
  showstringspaces=false,
  frame=single,
  breaklines=true,
  breakindent=0pt, 
  mathescape=true,
  lineskip=-1pt, 
}

\lstdefinestyle{pythonstyle}{
  numbers=none,
  numbersep=-6pt,
  backgroundcolor=\color{lightgray},
  basicstyle=\ttfamily\scriptsize,
  keywordstyle=\color{darkblue}\bfseries,
  commentstyle=\color{codegray}\itshape,
  stringstyle=\color{codegray},
  showstringspaces=false,
  frame=single,
  breaklines=true,
  breakindent=0pt,
  language=Python,
}

\usetikzlibrary{shapes.misc}

\crefname{figure}{Fig.}{Figures}

\usepackage{dutchcal}
\newtcolorbox{promptbox}[1][]{
  enhanced,
  colback=gray!8,
  colframe=black,
  boxrule=0.6pt,
  arc=3mm,
  left=3mm,right=3mm,top=2mm,bottom=2mm,
  drop shadow=black!30,
  title=#1,
  fonttitle=\bfseries,
}

\newcommand{\method}{{LM2LM}\xspace}

\newcommand{\coltablebox}[1]{\adjustbox{max width=\linewidth}{#1}}
\newlength{\colfigwidth}
\AtBeginDocument{\setlength{\colfigwidth}{0.465\textwidth}}

\title{\parbox{\linewidth}{\centering Language Models Learn to Improve Language Models}\vspace{.5em}}

\author[1]{Minhua Lin}
\author[2]{Zijun Wang}
\author[3]{Zewen Liu}
\author[4]{Zhaotian Weng}
\author[5]{Bing He}
\author[5]{Zhan Shi}
\author[5]{Benoit Dumoulin}
\author[1]{Xiang Zhang}
\author[1]{Suhang Wang}
\author[5]{Hanqing Lu}

\affiliation[1]{The Pennsylvania State University}
\affiliation[2]{UC Santa Cruz}
\affiliation[3]{Emory University}
\affiliation[4]{UC Santa Barbara}
\affiliation[5]{Amazon}

\abstract{\renewenvironment{abstract}{}{}\begin{abstract}
Large language models (LLMs) are increasingly used to automate AI research, including language-model (LM) development, offering a potential path toward recursive self-improvement.
However, turning experimental feedback into a learned ability to propose effective improvements across models faces two challenges.
First, training such an ability with real feedback is expensive: reinforcement learning (RL) requires evaluating tens of thousands of proposals, each potentially requiring a full model-training run.
Even at ten GPU-hours per run, reward collection alone would consume hundreds of thousands of GPU-hours, motivating a cheaper surrogate.
Second, surrogate feedback can become unreliable as policy optimization favors overestimated proposals when the proposal distribution shifts.
To address these challenges, we propose \method, which trains LLMs with RL to generate effective LM improvement proposals.
To provide scalable feedback, we use a research world model (RWM), a frozen LLM that predicts proposal gains from real experimental records provided in context, as the reward model for RL.
To maintain feedback reliability, we periodically execute current-policy proposals ranked highly by the RWM and add their measured outcomes to its context.
Across eight autoregressive and diffusion LM development environments, real-execution evaluations on held-out tasks show that \method enables Qwen3.8-27B to outperform GPT-6-sol, even when GPT-6-sol is given 300 real experimental records in context, while requiring 33\% lower data-generation cost.
\method also generalizes to a model family unseen during training.

\end{abstract}
}
\correspondence{\email{luhanqin@amazon.com}}

\begin{document}
\maketitle

\section{Introduction}\label{sec:intro}

Large language models (LLMs)~\cite{radford2018improving,radford2019language,touvron2023llama} are increasingly being adopted as agents for automated AI research (AutoResearch)~\cite{karpathy2026autoresearch}, where they propose ideas, implement changes, and execute experiments~\cite{si2026towards,novikov2025alphaevolve}. A promising application is language-model development (LM development), where agents explore architectural, optimization, and training choices and measure their effects on model performance~\cite{chi2026ai4ai,yang2026frontis}. These experiments generate feedback that could help LLMs learn how to improve language models. Developing this capability could accelerate model development and support progress toward recursive self-improvement (RSI)~\cite{chi2026ai4ai}~(Fig.~\ref{fig:overview}(a)).

Despite these advances, deciding what to try, i.e., \emph{research judgment}, remains a bottleneck for LLM-based research agents. Agents can implement and execute complex experiments, yet still struggle to identify changes that yield meaningful improvements~\cite{li2026genebench,si2026towards,chi2026ai4ai}. In LM development, effective choices depend on the target model and its training configuration: a change that benefits one setting may not benefit another. With expensive training experiments, selecting promising changes is essential to making effective use of limited experimental resources. We therefore ask: \textit{how can language models learn to make better model-improvement decisions and transfer this capability to new models and training configurations?}

A natural way to improve research judgment is to learn from past experiments. Experimental records can inform an LLM's decisions through in-context learning (ICL)~\cite{brown2020language,novikov2025alphaevolve,si2026towards}, and can also improve predictions of experimental outcomes and the selection of promising proposals~\cite{wang2026language}. This suggests using experimental knowledge as feedback for parameter learning to improve an LLM's proposal-generation capability beyond in-context learning. However, learning from this feedback raises two key challenges (Fig.~\ref{fig:overview}(b)-(c)).


\begin{frameworkfigure}
    \centering
    \includegraphics[width=\linewidth]{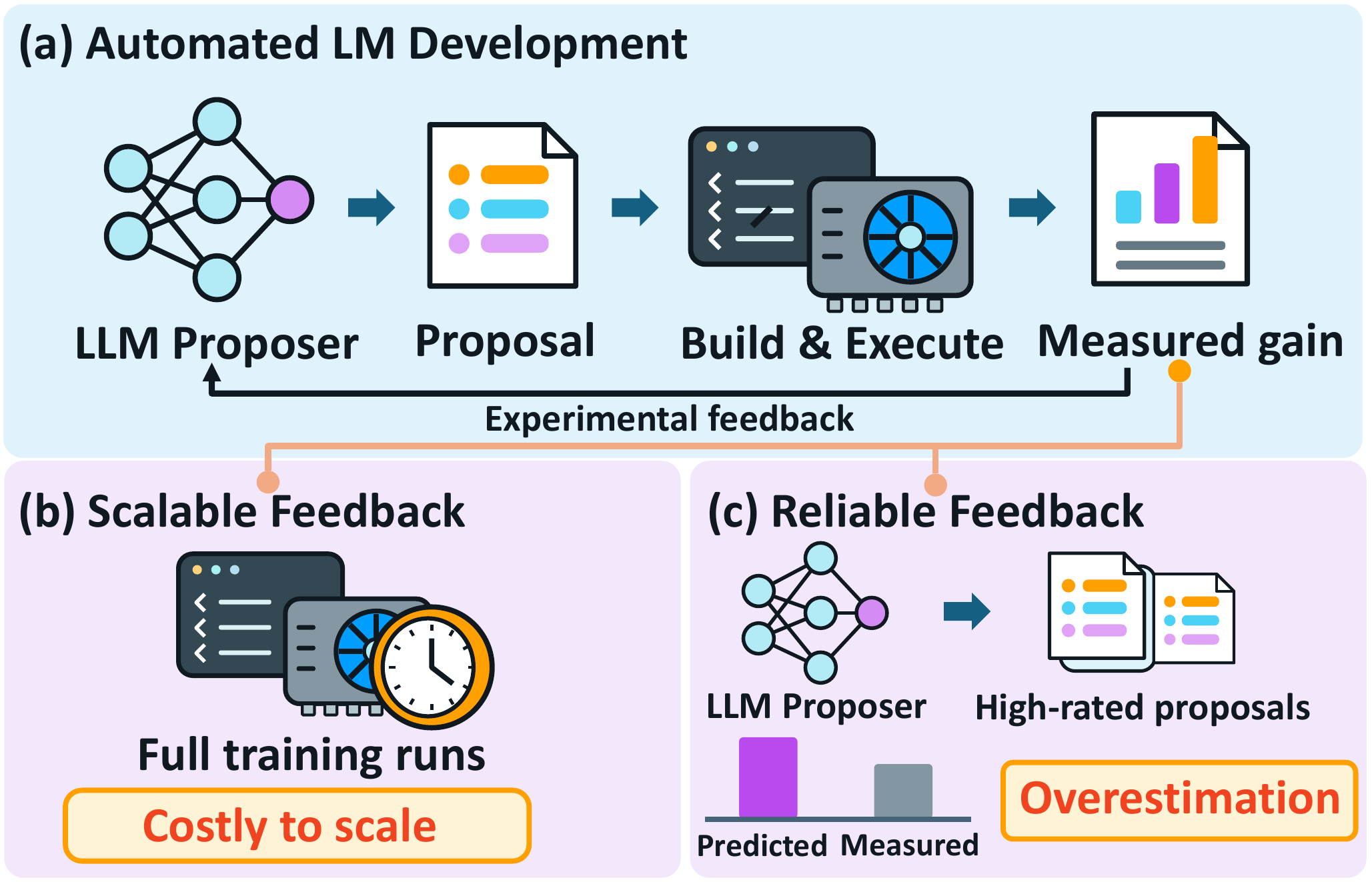}
    \vspace{-1.5em}
    \caption{\textbf{Overview of automated LM development.}}
    \label{fig:overview}
\end{frameworkfigure}

First, \emph{how can we obtain cheap and scalable training feedback?}
Training LLMs with reinforcement learning (RL) to generate better proposals requires feedback on numerous sampled proposals.
The most direct feedback comes from executing each proposal and measuring its outcome~\cite{si2026towards}, but this requires costly model-training runs.
For example, training a 486M-parameter model on 8B tokens takes approximately 45 B200 GPU-hours.
A GRPO-style RL~\cite{shao2024deepseekmath} setup with 100 updates, 32 training states per update, and 8 proposals per state would require 25,600 proposal evaluations. Executing these proposals at the same per-run cost would consume approximately 1,152,000 GPU-hours, costing over \$16 million.
This motivates low-cost surrogate feedback without executing every proposal.
Second, \emph{how can we keep surrogate feedback reliable as learning progresses?} Accurate predictions on past experiments do not ensure reliable feedback throughout training. As learning changes which proposals the LLM generates, the surrogate must evaluate proposals that its experimental records may not cover. RL may then favor overestimated proposals, increasing surrogate rewards without corresponding real gains~\cite{gao2023scaling,zhang2026chasing}. A scalable learning approach must therefore keep its feedback grounded in real outcomes on the proposals the current LLM generates.



To address these challenges, we propose \method, a framework that trains LLMs with RL to generate effective proposals for LM development.
To obtain cheap and scalable training signals, we use a frozen LLM as a research world model (RWM)~\cite{wang2026language} to predict experimental gains relative to a reference configuration without executing proposals.
Grounded in an initial set of real experimental records provided as in-context knowledge, the RWM supplies predicted gains for constructing RL rewards~\cite{yu2026dapo}.
To keep this feedback reliable as training progresses, we introduce \emph{on-policy calibration}: we periodically execute a subset of the current policy's proposals that the RWM ranks highly and incorporate their measured outcomes into its in-context knowledge. This targets real execution at proposals favored by RL, where overestimation can misdirect learning, and helps maintain prediction reliability as LLMs' proposals change.

Our contributions are:
(i) \textbf{Method.} We propose \method, which trains LLMs with RL using RWM-based rewards grounded in real records and calibrated through targeted real execution;
(ii) \textbf{Experiments.} Extensive real-execution experiments on held-out environments show that \method enables Qwen3.8-27B to outperform frontier models such as GPT-6-sol and surpass baselines leveraging past experimental records at lower data-generation cost;
(iii) \textbf{Analysis.} We show that RWM-based feedback provides informative rankings of proposal gains, while on-policy calibration mitigates overestimation and largely preserves ranking quality as training progresses, unlike a fixed-record RWM that eventually performs worse than random.

\section{Related Work}\label{sec:related}
\noindent\textbf{AutoResearch with LLM Agents.}
AutoResearch aims to automate empirical AI research through iterative idea generation, experimental implementation, execution, and analysis.
Early work shows LLM agents' ability to generate research ideas and conduct end-to-end experiments~\cite{baek2025researchagent,si2025can,lu2024ai,weng2026deepscientist}, with benchmarks evaluating these capabilities in machine-learning engineering and AI R\&D~\cite{huang2024mlagentbench,chan2025mle,wijk2025rebench,chi2026ai4ai}. To improve research outcomes, some methods iterate on candidate solutions with execution feedback while keeping the underlying agent fixed~\cite{jiang2025aide,novikov2025alphaevolve,karpathy2026autoresearch}. Others instead train the agent on execution feedback, for machine-learning engineering~\cite{yang2026frontis} and LLM model development~\cite{si2026towards}.
However, these methods rely on costly real execution to obtain training signals, limiting learning at scale. In contrast, we train LLMs with feedback from a world model that predicts experimental outcomes, enabling scalable learning with fewer real executions. 
Full details of related works are in Appendix~\ref{app:related}.

\section{Methodology}\label{sec:method}
\begin{frameworkfigure}[15]
    \centering
    \includegraphics[width=\linewidth]{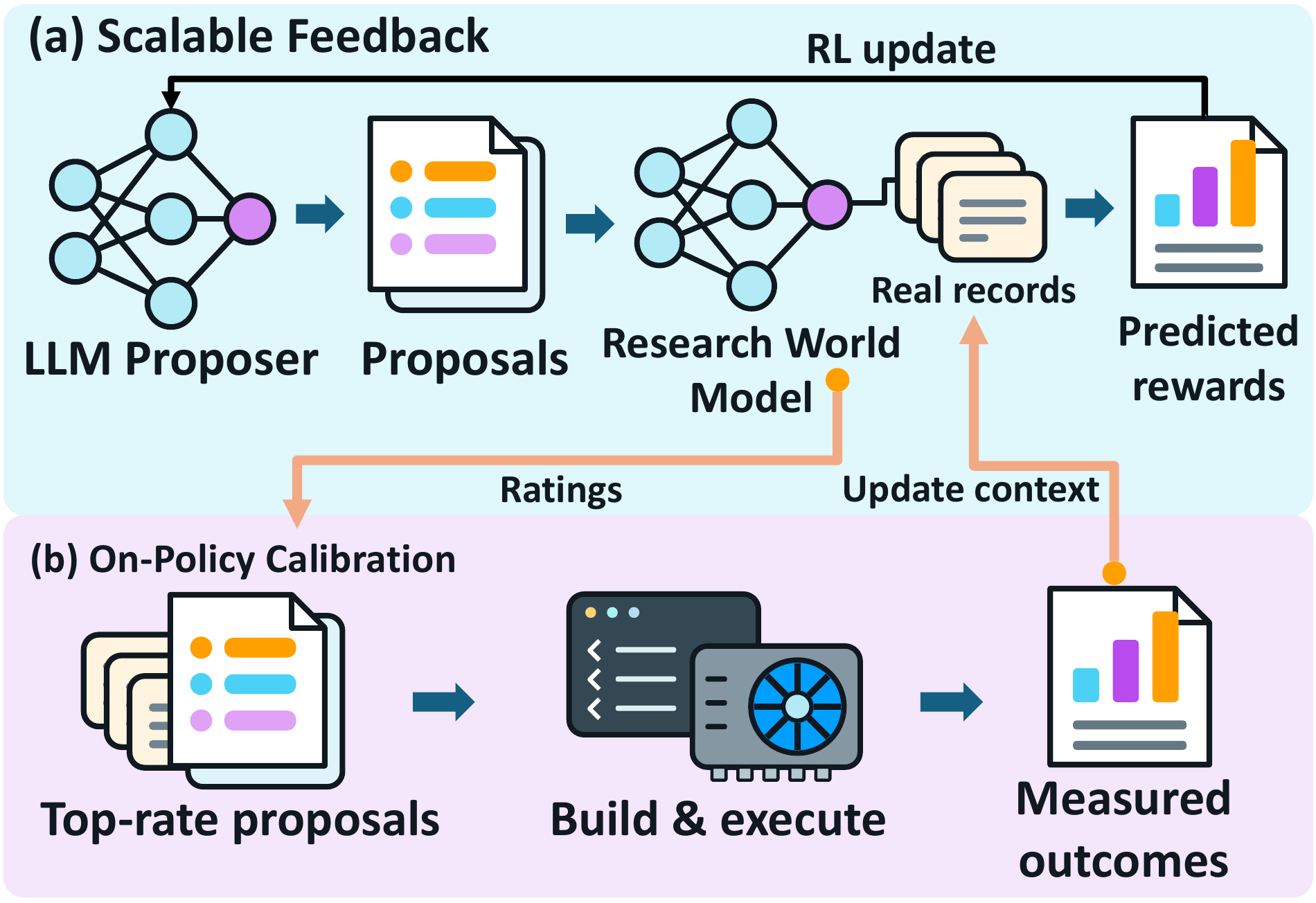}
    \vspace{-1.5em}
    \caption{\textbf{Overview of \method.}}
    \label{fig:framework}
\end{frameworkfigure}

In this section, we present \method.
We first formalize AutoResearch for LM development
and motivate RL-based proposer training
(Sec.~\ref{ssec:formulation}).
To provide scalable feedback, we use a frozen LLM
as a research world model (RWM) to predict
experimental gains without executing proposals,
and use its predictions as RL rewards
(Sec.~\ref{ssec:rwm-feedback}).
To maintain reliable feedback as the proposer evolves,
we periodically execute high-rated proposals from
the current policy and incorporate their measured
outcomes into the RWM's context
(Sec.~\ref{ssec:rwm-reliability}).
An overview of \method is shown in
Fig.~\ref{fig:framework}.

\subsection{Preliminaries}\label{ssec:formulation}

\noindent\textbf{AutoResearch for LM Development.}
We consider a set of AutoResearch tasks $\mathcal E$, where each task $e\in\mathcal E$ seeks to improve a target language model by modifying its architecture, training objective, or optimization. Each task specifies a model family, model size, pretraining budget, and a reference configuration with validation loss $b_e$.



A research campaign on task $e$ involves two LLM-based roles: a \emph{proposer} $\pi_\theta$, which generates proposals, and a \emph{builder} $\mathcal B$, which translates them into executable modifications to the reference configuration. A campaign proceeds through repeated proposal-and-execution steps.
At step $t$, the proposer observes the research state $s_t=(e,h_t)$, where $h_t$ records the proposals executed in previous steps and their measured outcomes, and generates a proposal $a_t\sim\pi_\theta(\cdot\mid s_t)$.
The builder implements $a_t$, and the resulting training run yields validation loss $\ell_t$. The experimental gain is defined as
\begin{equation}
    g(s_t,a_t)=b_e-\ell_t,
    \label{eq:real-gain}
\end{equation}
where $g(s_t,a_t)>0$ indicates an improvement over the reference configuration.
The proposal and its measured outcome are appended to the history, producing
$h_{t+1}$ and the next research state $s_{t+1}=(e,h_{t+1})$.


\noindent\textbf{Problem Statement.}
An AutoResearch campaign consists of a sequence of research steps, where each step requires the proposer to decide what proposal to explore next based on the current task and previous experimental outcomes. In this paper, our goal is then to improve this decision-making capability by training a proposer $\pi_\theta$ to generate proposals with high experimental gain at each step.

Formally, let $\mathcal E_{\mathrm{train}}$ and
$\mathcal E_{\mathrm{test}}$ denote disjoint sets
of training and held-out tasks, respectively,
and let $\mu_{\mathrm{train}}$ and
$\mu_{\mathrm{test}}$ denote the corresponding
distributions over research states $s=(e,h)$. For $\nu\in\{\mathrm{train},\mathrm{test}\}$, we define the expected experimental gain as 
\begin{equation}
    J_{\nu}(\theta)
    =
    \mathbb E_{
        s\sim\mu_{\nu},\,
        a\sim\pi_\theta(\cdot\mid s)}
    \left[g(s,a)\right],
    \label{eq:objective}
\end{equation} 
where $J_{\nu}(\theta)$ measures the expected experimental gain of proposals generated by $\pi_\theta$ across research states sampled from $\mu_{\nu}$. We seek to maximize $J_{\mathrm{train}}(\theta)$ using training tasks, with the aim of achieving high $J_{\mathrm{test}}(\theta)$ on held-out tasks.

\noindent\textbf{RL with Experimental Feedback.}
Since the experimental gain $g(s,a)$ directly measures the quality of a proposal, it can serve as a verifiable reward, making reinforcement learning with verifiable rewards (RLVR) a natural approach to optimizing $J_{\mathrm{train}}(\theta)$~\cite{lambert2025tulu}.
A representative algorithm is DAPO~\cite{yu2026dapo}: for a state $s\sim\mu_{\mathrm{train}}$, it samples a group of $G$ proposals $\{a_i\}_{i=1}^{G}$ from the current policy $\pi_{\theta_{\mathrm{old}}}(\cdot\mid s)$, scores them with rewards $\{R_i\}_{i=1}^{G}$, and maximizes
\begin{equation}
\begin{aligned}
    \mathcal J_{\mathrm{DAPO}}(\theta)
    = \mathbb{E}\Big[&\tfrac{1}{\sum_{i}|a_i|}\textstyle\sum_{i=1}^{G}\sum_{k=1}^{|a_i|}\\
    &\min\big(\rho_{i,k}\hat A_i,\,\bar\rho_{i,k}\hat A_i\big)\Big],
\end{aligned}
    \label{eq:dapo}
\end{equation}
where $|a_i|$ is the number of tokens in $a_i$, $\rho_{i,k}=\pi_\theta(a_{i,k}\mid s,a_{i,<k})/ \pi_{\theta_{\mathrm{old}}}(a_{i,k}\mid s,a_{i,<k})$ is the token-level probability ratio, $\bar\rho_{i,k}=\mathrm{clip}(\rho_{i,k}, 1-\epsilon_{\mathrm{low}},1+\epsilon_{\mathrm{high}})$, and the advantage $\hat A_i=\big(R_i-\mathrm{mean}(\{R_j\}_{j=1}^{G})\big)/\mathrm{std}(\{R_j\}_{j=1}^{G})$. Groups with no reward variation are discarded.

\subsection{Scalable Feedback: Research World Model as the Verifier}\label{ssec:rwm-feedback}
RLVR typically relies on inexpensive verification,
e.g., checking the correctness of a mathematical
answer~\cite{cobbe2021training}.
However, directly applying RLVR to AutoResearch
for LM development is computationally expensive,
as evaluating each proposal requires training
the target model to measure its experimental gain.
Since RL samples numerous proposals across research
states and training updates, obtaining execution-based
rewards at scale becomes prohibitively costly.
To address this challenge, we use a research world
model (RWM) to predict experimental gains without
executing every proposal, and train the proposer with
DAPO on rewards built from these predictions.

\noindent\textbf{Research World Model.}
An RWM predicts the outcomes of AI research experiments without executing them~\cite{wang2026language}. We instantiate it as a frozen LLM $W$. To improve prediction reliability, we ground it in real experimental records. Specifically, 
we collect $N_0$ experimental records from training tasks to form the initial record set $M_0$. Each record $(e,a,g)$ contains a task, an executed proposal, and its measured gain. Formally, given a task $e$, proposal $a$, and experimental records $M$, the RWM predicts the experimental gain $\hat g(e,a;M)$ as
\begin{equation}
    \hat g(e,a;M) = W(e,a;M),
    \label{eq:rwm}
\end{equation}
where $M$ is provided as in-context evidence.
Thus, evaluating a proposal requires only an LLM call rather than a model-training run.

\noindent\textbf{Reward Design.}
A straightforward approach is to use RWM-predicted gains directly as RL rewards. This poses two challenges. First, predicted gains vary in scale across tasks and may contain extreme values, allowing tasks with larger gain scales or a few extreme predictions to dominate policy optimization. Second, high predicted gains do not necessarily imply useful proposals, as proposals may be invalid, repetitive, or insufficiently diverse.

To address the first challenge, we normalize predicted gains by a task-specific scale $\tau_e$ and clip them at a threshold $c$:
\begin{equation}
    r(s,a;M) =
    \mathrm{clip}\big(
        \hat g(e,a;M)/\tau_e,\,-c,\,c
    \big),
    \label{eq:reward}
\end{equation}
where $e$ is the task associated with research state $s$, and $\tau_e$ is determined by reference seed noise and the dispersion of RWM-predicted gains.
Normalization makes rewards comparable across tasks, while clipping limits the influence of extreme predictions on policy updates.

To address the second challenge, we incorporate proposal validity, novelty, and diversity into the final rewards.
For a group of $G$ proposals sampled from state $s$, a proposal is \emph{invalid} if it cannot be implemented by the builder, \emph{repeated} if it reproduces the reference configuration or a proposal in the history $h$, and \emph{valid} otherwise.
Let $\mathcal V$ denote the set of valid proposals within the group.
We assign
\begin{equation}
    R_i =
    \begin{cases}
        r(s,a_i;M)+b_i, & a_i\in\mathcal V,\\[2pt]
        \min_{j\in\mathcal V}R_j-\delta, & a_i \text{ repeated},\\[2pt]
        -c, & a_i \text{ invalid},
    \end{cases}
    \label{eq:rl-reward}
\end{equation}
where $\delta>0$ is a repetition margin and $b_i=\lambda\sigma_G\tilde d_i$ is a diversity bonus.
Here, $d_i=1/n_i$, where $n_i$ is the number of valid proposals sharing the same mechanism direction as $a_i$ within the group; $\tilde d_i$ is the within-group standardized value of $d_i$, $\sigma_G$ is the standard deviation of normalized predicted rewards within the group, and $\lambda$ controls the bonus strength.
The diversity bonus favors less common mechanism directions to encourage broader exploration during RL.
Repeated proposals provide no new experimental information but may receive high predicted gains, so $\delta$ ranks them below all valid proposals. More reward details are in Appendix~\ref{app:impl-reward}.

\noindent\textbf{Policy Optimization.}
With the rewards in Eq.~\eqref{eq:rl-reward}, we train the proposer using DAPO (Eq.~\eqref{eq:dapo}). Since rewards are already normalized by $\tau_e$ to be comparable across tasks, we use group-mean-centered advantages $\hat A_i=R_i-\mathrm{mean}(\{R_j\}_{j=1}^{G})$ without dividing by the group's standard deviation~\cite{liu2025understanding}, preserving reward dispersion across groups.


Beyond providing RL rewards, the RWM also enables low-cost expansion of RL training states.
Each RL training state $s=(e,h)$ pairs a training task $e$ with a history $h$ sampled from $\{(e',a,g)\in\mathcal P : e'=e\}$, where the record pool $\mathcal P$ is initialized as $M_0$.
Since $M_0$ contains only a few records per task and scaling up $\mathcal P$ with real outcomes requires costly execution, we periodically add the state--proposal pairs $\mathcal D$ sampled during RL to $\mathcal P$, with their RWM-predicted gains as synthetic outcomes (Appendix~\ref{app:impl-refresh}):
\begin{equation}
    \mathcal P\leftarrow\mathcal P\cup\big\{\big(e,a,\hat g(e,a;M)\big) : \big((e,h),a\big)\in\mathcal D\big\}.
    \label{eq:pool}
\end{equation}
Thus, both RWM-based rewards and synthetic training states enable scalable RL without executing every sampled proposal.

\subsection{Reliable Feedback: On-Policy Calibration of the RWM}
\label{ssec:rwm-reliability}
\noindent\textbf{Prediction Bias under Policy Shift.}
While the RWM provides scalable feedback, RL increasingly favors highly rated proposals that may not be adequately covered by the initial records $M_0$, making it susceptible to reward overoptimization~\cite{gao2023scaling,zhang2026chasing}.
To characterize this risk, we decompose the normalized real objective as
\begin{equation}
    \bar J(\theta)
    =\underbrace{\mathbb E\big[r(s,a;M)\big]}_{\hat J_M(\theta)}
    +\underbrace{\mathbb E\big[\bar g(s,a)-r(s,a;M)\big]}_{\Delta_M(\theta)},
    \label{eq:bias}
\end{equation}
where $\bar J(\theta)=\mathbb E[\bar g(s,a)]$, $\bar g(s,a)=\mathrm{clip}(g(s,a)/\tau_e,-c,c)$, and the expectations are over $s\sim\mu_{\mathrm{train}}$ and $a\sim\pi_\theta(\cdot\mid s)$.
Here, $\hat J_M(\theta)$ is the expected normalized predicted gain, and $\Delta_M(\theta)$ is its prediction bias under the current proposer.
As RL optimizes RWM-based rewards without observing this bias, $\hat J_M$ may increase without a corresponding increase in $\bar J$.

\noindent\textbf{Targeted Calibration with Real Execution.}
To mitigate this risk, we introduce on-policy calibration: rather than keeping the RWM's records fixed, \method periodically executes high-rated proposals from the current proposer and incorporates their real outcomes into the records to improve RWM predictions as the proposer evolves.

Specifically, at each update point $u$, let $\mathcal C_u$ denote the current proposer's samples on training states, excluding proposals whose outcomes are already in $M_u$.
We select $B_u$ high-rated proposals:
\begin{equation}
    \mathcal Q_u
    =
    \operatorname*{arg\,max}_{\mathcal Q\subseteq\mathcal C_u,\;|\mathcal Q|=B_u}
    \;\sum_{(s,a)\in\mathcal Q} r(s,a;M_u).
    \label{eq:select}
\end{equation}
The builder implements and executes these proposals, and their measured outcomes are added to the records:
\begin{equation}
    M_{u+1} = M_u \cup \big\{(e,a,g(s,a)) : (s,a)\in\mathcal Q_u\big\},
    \label{eq:update}
\end{equation}
where $e$ is the task of $s$. Subsequent RL steps compute rewards with $W(\cdot\,;M_{u+1})$. Calibration occurs between RL training segments, keeping the RWM feedback fixed within each segment.


Since calibration updates the RWM through in-context records without retraining, it requires only $\sum_{u=0}^{U-1}B_u$ additional real executions, independent of the number of proposals sampled during RL. The training algorithm is summarized in Alg.~\ref{alg:method}.
\section{Experiments}\label{sec:exp}



In this section, we first compare \method with baselines on proposal quality and data-generation cost (Sec.~\ref{ssec:main}) and evaluate its generalization to an unseen model family (Sec.~\ref{ssec:RQ3}). We then conduct ablation studies to understand the contribution of on-policy calibration (Sec.~\ref{ssec:ablation}) and analyze why RWM feedback can guide RL and how calibration sustains it (Sec.~\ref{ssec:RQ2}).

\subsection{Experimental Setup}\label{ssec:exp-env}
\noindent\textbf{Environments.} 
We construct language-model training environments spanning four autoregressive (AR) families: OLMo-3~\cite{olmo2025olmo}, Qwen3~\cite{yang2025qwen3}, NanoChat~\cite{nanochat}, and Marin~\cite{hall2025marin}; and four diffusion families: \mbox{E2D2-SW}, E2D2~\cite{arriola2026encoder}, MDM~\cite{nie2025scaling}, and BD3LM~\cite{arriola2025block}.
Each task specifies a model family, model size, training budget, and reference training recipe.
Each experiment trains the target model from scratch.
Further details are in Appendix~\ref{app:env}.

\noindent\textbf{Data Construction.} We split the tasks into 40 training and 10 held-out tasks.
We curate and execute 300 proposals from a library of training-recipe mechanisms on the training tasks, yielding 300 experimental records.
Of these, 100 serve as the initial records $M_0$ of \method, grounding the RWM.
We construct initial RL training states $s=(e,h)$ with histories $h$ containing 0--8 records of task $e$ from $M_0$, expanded during RL (Eq.~\eqref{eq:pool}).
For evaluation, we construct 60 AR and 18 diffusion held-out research states using only real experimental histories. More details are in Appendix~\ref{app:data-construction}.


\begin{table*}[t]
\centering
\small
\setlength{\tabcolsep}{3pt}
\caption{\textbf{Comparison with baselines on held-out research states.}
Experimental gains are reported in $10^{-3}$ BPB (AR) and $10^{-2}$ nats per token (diffusion).
Best results in Avg@4, Best@4, Neg., and Row vs.\ Base are in \textbf{bold}.}
\vspace{-0.5em}
\label{tab:main}
\resizebox{0.7\textwidth}{!}{
\begin{tabular}{l|ccc|cc}
\toprule
\multirow{2}{*}{\textbf{Model}}
& \multicolumn{3}{c|}{\textbf{Experimental Gain}}
& \multicolumn{2}{c}{\textbf{Pairwise Win Rate} (\%) $\uparrow$} \\
\cmidrule(lr){2-4}\cmidrule(lr){5-6}
& \textbf{Avg@4} $\uparrow$ & \textbf{Best@4} $\uparrow$ & \textbf{Neg.} (\%) $\downarrow$
& \textbf{Row vs.\ Base} & \textbf{\method\ vs.\ Row} \\
\midrule
\multicolumn{6}{@{}l}{\emph{AR families}} \\
Base & $-7.94$ & $11.39$ & $70.4$ & -- & $93.3$ \\
Base-ICL & $4.04$ & $16.54$ & $40.8$ & $81.7$ & $66.7$ \\
GPT-6-sol & $6.51$ & $8.79$ & $52.5$ & $83.3$ & $56.7$ \\
GPT-6-sol-ICL & $6.15$ & $11.27$ & $\mathbf{29.6}$ & $78.3$ & $51.7$ \\
SFT & $-12.19$ & $5.51$ & $71.5$ & $55.0$ & $93.3$ \\
RFT-Real & $-22.95$ & $9.69$ & $66.5$ & $45.0$ & $95.0$ \\
\method & $\mathbf{17.06}$ & $\mathbf{23.40}$ & $39.2$ & $\mathbf{93.3}$ & -- \\
\midrule
\multicolumn{6}{@{}l}{\emph{Diffusion families}} \\
Base & $-4.18$ & $5.89$ & $59.7$ & -- & $100.0$ \\
Base-ICL & $4.03$ & $8.50$ & $13.9$ & $83.3$ & $72.2$ \\
GPT-6-sol & $1.98$ & $4.43$ & $38.9$ & $83.3$ & $83.3$ \\
GPT-6-sol-ICL & $4.94$ & $5.93$ & $\mathbf{0.0}$ & $94.4$ & $72.2$ \\
SFT & $-3.09$ & $3.73$ & $54.2$ & $61.1$ & $100.0$ \\
RFT-Real & $-3.72$ & $3.37$ & $52.8$ & $50.0$ & $88.9$ \\
\method & $\mathbf{7.10}$ & $\mathbf{9.93}$ & $7.6$ & $\mathbf{100.0}$ & -- \\
\bottomrule
    \end{tabular}}
\end{table*}

\begin{figure*}[t]
\centering
\begin{subfigure}[t]{0.24\textwidth}
\centering
\includegraphics[width=\linewidth]{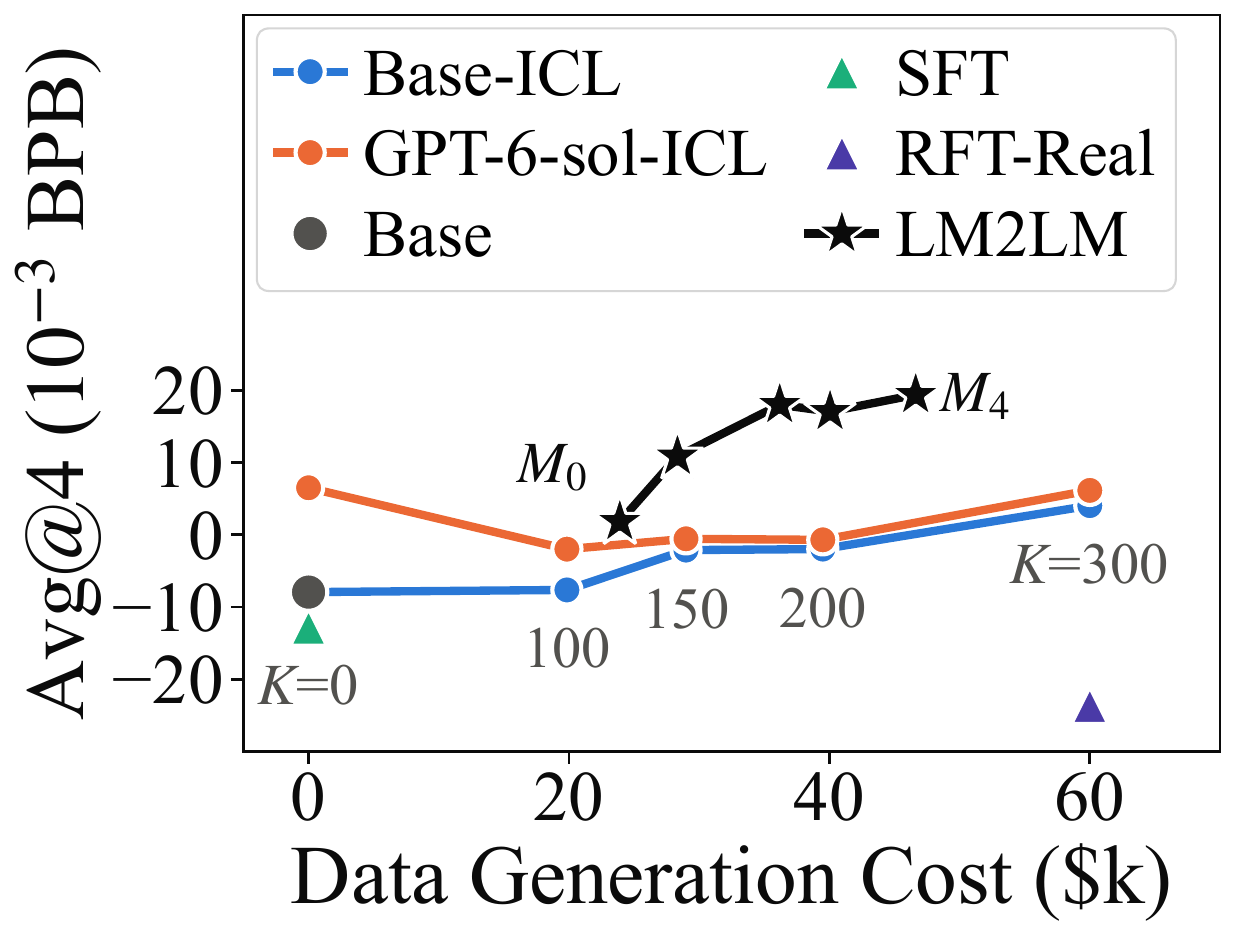}
\caption{AR: Avg@4 vs.\ cost.}
\label{fig:cost-ar-cost}
\end{subfigure}\hfill
\begin{subfigure}[t]{0.24\textwidth}
\centering
\includegraphics[width=\linewidth]{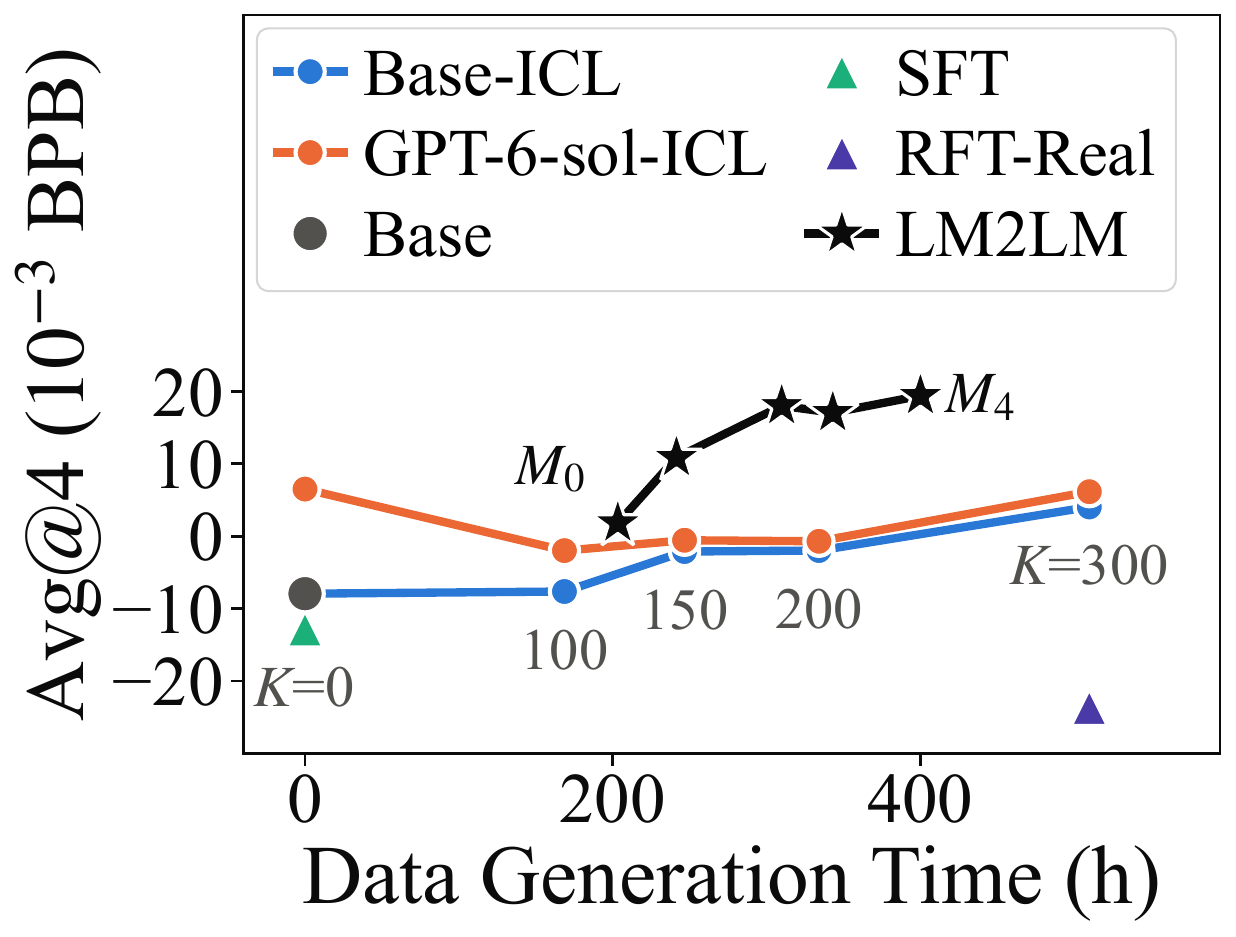}
\caption{AR: Avg@4 vs.\ time.}
\label{fig:cost-ar-time}
\end{subfigure}\hfill
\begin{subfigure}[t]{0.24\textwidth}
\centering
\includegraphics[width=\linewidth]{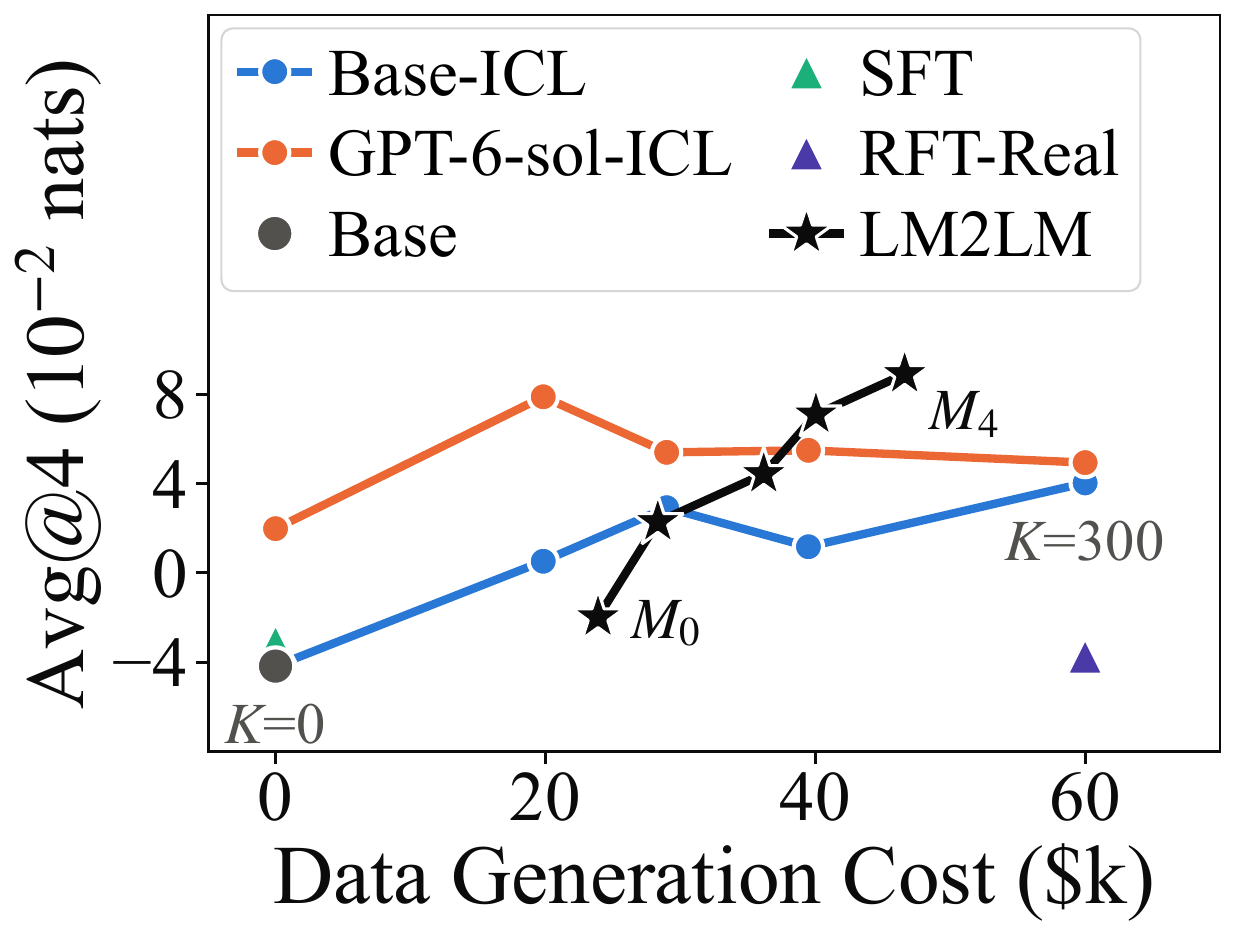}
\caption{Diff.: Avg@4 vs.\ cost.}
\label{fig:cost-diff-cost}
\end{subfigure}\hfill
\begin{subfigure}[t]{0.24\textwidth}
\centering
\includegraphics[width=\linewidth]{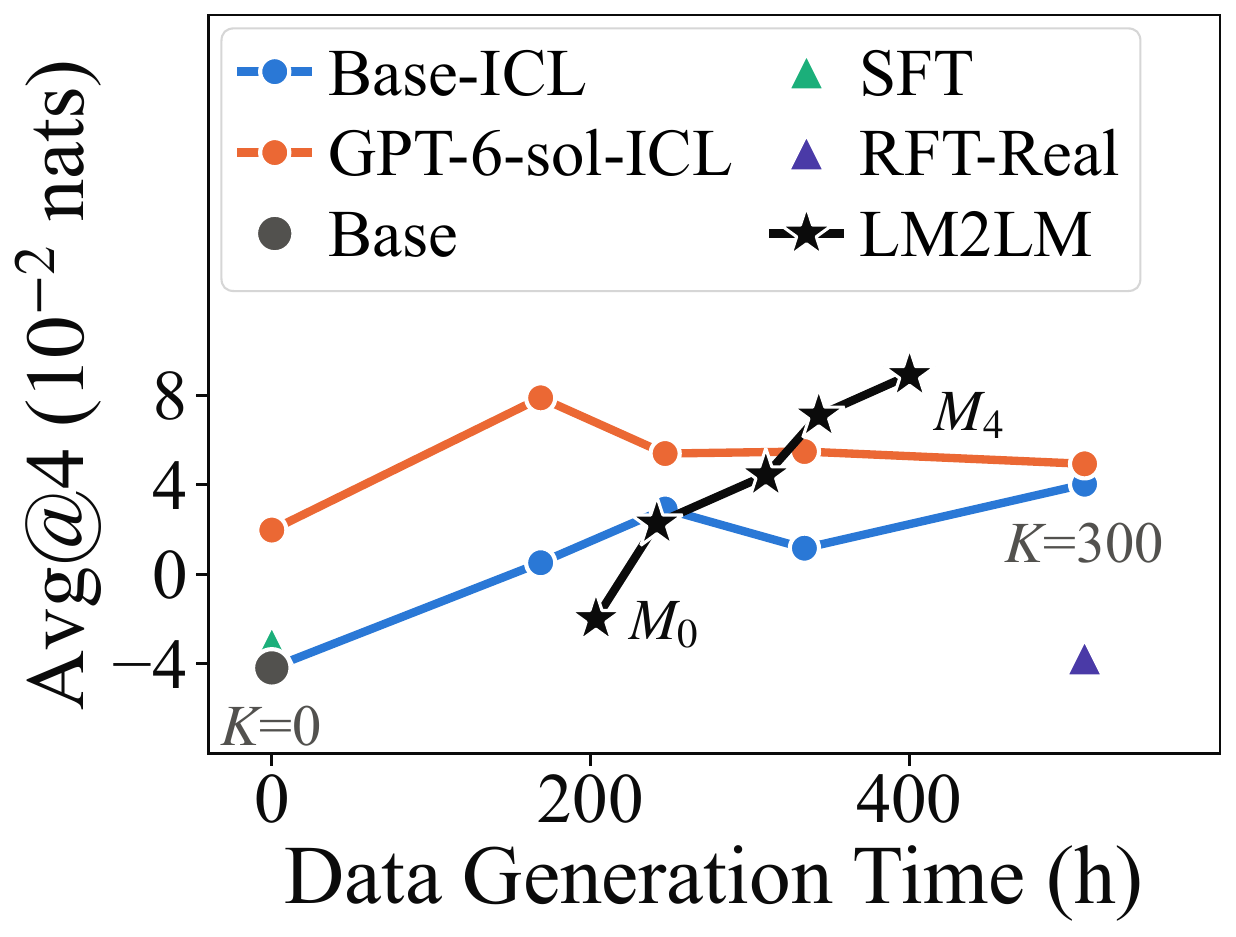}
\caption{Diff.: Avg@4 vs.\ time.}
\label{fig:cost-diff-time}
\end{subfigure}
\caption{Real experimental gains (Avg@4) versus data-generation cost and time on held-out research states.
(a, b) show AR tasks in $10^{-3}$ BPB, while (c, d) show diffusion tasks in $10^{-2}$ nats per token.}
\vspace{-0.5em}
\label{fig:cost}
\end{figure*}

\noindent\textbf{Baselines.}
We compare \method with training-free and fine-tuning-based methods.
(i) \emph{Training-free proposers}:
We evaluate \method's base model and a frontier LLM (GPT-6-sol) in both zero-shot and in-context learning (ICL) settings.
For ICL, we use the above 300 experimental records as in-context examples.
(ii) \emph{Fine-tuned proposers}:
SFT fine-tunes the base model on 300 (state, proposal) pairs constructed from the same records, and RFT-Real applies rejection-sampling fine-tuning~\cite{yuan2023scaling}, keeping only the pairs whose proposals have positive measured gains.
Both use the same base model and builder as \method.

\noindent\textbf{Evaluation Protocol.}
We evaluate proposers on held-out research states through real execution, with each proposer generating four proposals per state.
We measure experimental gains using validation bits per byte (BPB) for AR models and negative log-likelihood (NLL) for diffusion models.
We report Avg@4, Best@4, negative-gain rate (Neg.), and pairwise win rates based on Avg@4.
To assess efficiency, we measure data-generation cost and time, including real execution and RWM prediction costs. GPU time is priced at \$14.208 per B200 GPU-hour and RWM calls at GPT-6-sol token rates.
More details are in Appendix~\ref{app:eval-prot}.

\noindent\textbf{Implementation Details.}
We use Qwen3.8-27B~\cite{qwen38} as the proposer backbone, GPT-6-sol~\cite{openai2026gpt6sol} as the RWM, and GPT-6-astra~\cite{openai2026gpt6astra} as the builder, both at high reasoning effort.
We run DAPO for 96 steps, refreshing rollouts every 8 steps with 256 training states and $G=8$ proposals per state, and clipping rewards at $c=6$.
Starting with $N_0=100$ initial experimental records, we calibrate the RWM three times ($U=3$), after steps 16, 40, and 64, by executing proposals selected using Eq.~\eqref{eq:select}, so that \method uses 200 real runs by default and each of the four training segments has a fixed RWM reward function.
Unless otherwise specified, we report results at step 96.
More details are in Appendix~\ref{sec:exp-impl-details}.

\subsection{Main Comparison with Baselines}\label{ssec:main}
\noindent\textbf{Main Results.}
We compare \method with baselines on proposal quality across held-out research states, following the evaluation protocol in Sec.~\ref{ssec:exp-env}. Table~\ref{tab:main} reports the results. We observe: 
(i) \emph{Past experimental experience helps, but how it is used matters.}
With 300 experimental records, Base-ICL improves Avg@4 over Base from $-7.94$ to $4.04$ on AR tasks and from $-4.18$ to $4.03$ on diffusion tasks, demonstrating the value of past experimental experience. 
In contrast, SFT and RFT-Real fail to consistently improve proposal quality, suggesting that directly fine-tuning on a few hundred past proposals is insufficient to learn effective proposal generation.
(ii) \emph{\method outperforms all baselines.}
It achieves the highest Avg@4 and Best@4 on both AR (17.06/23.40) and diffusion (7.10/9.93) tasks, outperforming Base on 93.3\% and 100\% of states, respectively. This validates \method's effectiveness.
(iii) \emph{RL reduces harmful proposals while enabling larger gains.}
Compared with Base, \method reduces negative-gain proposals from 70.4\% to 39.2\% on AR tasks and from 59.7\% to 7.6\% on diffusion tasks.
Despite a higher negative-gain rate than GPT-6-sol-ICL on AR tasks (39.2\% vs.\ 29.6\%), \method achieves substantially higher Avg@4 (17.06 vs.\ 6.15), suggesting that it explores higher-gain proposals rather than simply avoiding negative outcomes.


\noindent\textbf{Cost Efficiency.}
We compare all methods in terms of Avg@4 and Best@4 versus data-generation cost and time. For ICL baselines, we vary the number of in-context records $K\in\{0,100,150,200,300\}$. For \method, we evaluate checkpoints at steps 16, 40, 64, 96, and 128, corresponding to RWM records $M_0$ through $M_4$.
Figure~\ref{fig:cost} reports Avg@4 results, with Best@4 results in Appendix~\ref{app:data-cost}.
We observe that 
(i) \emph{\method achieves larger gains with fewer real executions.}
At step 96 ($M_3$), \method achieves Avg@4 of 17.06 and 7.10 on AR and diffusion tasks, respectively, outperforming ICL with $K=200$ at comparable data-generation cost (\$40.1k vs.\ \$39.5k). Compared with ICL at $K=300$, it achieves higher gains with only 200 real runs, reducing data-generation cost and time by approximately 33\% (\$40.1k vs.\ \$60.0k; 343 vs.\ 510 hours).
Even including RL training, its total cost remains lower than ICL with $K=300$ (\$56.1k vs.\ \$60.0k), as detailed in Appendix~\ref{app:total-cost}.
(ii) \emph{\method improves further at limited additional cost as training progresses.}
Between steps 40 and 96, \method improves Avg@4 from 10.85 to 17.06 on AR tasks and from 2.28 to 7.10 on diffusion tasks. This requires only \$11.8k in additional data-generation cost, providing RWM rewards for over 14k sampled proposals with just 69 additional calibration records. These results show how \method scales training feedback with limited real execution.

\begin{table}[t]
\centering
\setlength{\tabcolsep}{3pt}
\caption{\textbf{Generalization to the held-out Qwen3 family.} Gains are reported in $10^{-3}$ BPB.
Best results in Avg@4, Best@4, and Row vs.\ Base are in \textbf{bold}.}
\label{tab:rq3-transfer-l2}
\coltablebox{
\begin{tabular}{l|cc|cc}
\toprule
\multirow{2}{*}{\textbf{Model}}
& \multicolumn{2}{c|}{\textbf{Gain} $\uparrow$}
& \multicolumn{2}{c}{\textbf{ Win Rate} (\%) $\uparrow$} \\
\cmidrule(lr){2-3}\cmidrule(lr){4-5}
& \textbf{Avg@4} & \textbf{Best@4}
& \textbf{Row vs.\ Base} & \textbf{\method\ vs.\ Row} \\
\midrule
Base          & $3.53$ & $31.39$ & -- & $87.5$ \\
Base-ICL      & $2.97$ & $25.22$ & $58.3$ & $100.0$ \\
GPT-6-sol     & $20.19$ & $22.75$ & $66.7$ & $62.5$ \\
GPT-6-sol-ICL & $21.63$ & $25.27$ & $75.0$ & $62.5$ \\
\method & $\mathbf{53.72}$ & $\mathbf{86.69}$ & $\mathbf{87.5}$ & -- \\
\bottomrule
\end{tabular}}
\end{table}

\begin{figure}[t]
\centering
\begin{subfigure}[t]{0.49\colfigwidth}\centering\includegraphics[width=\linewidth]{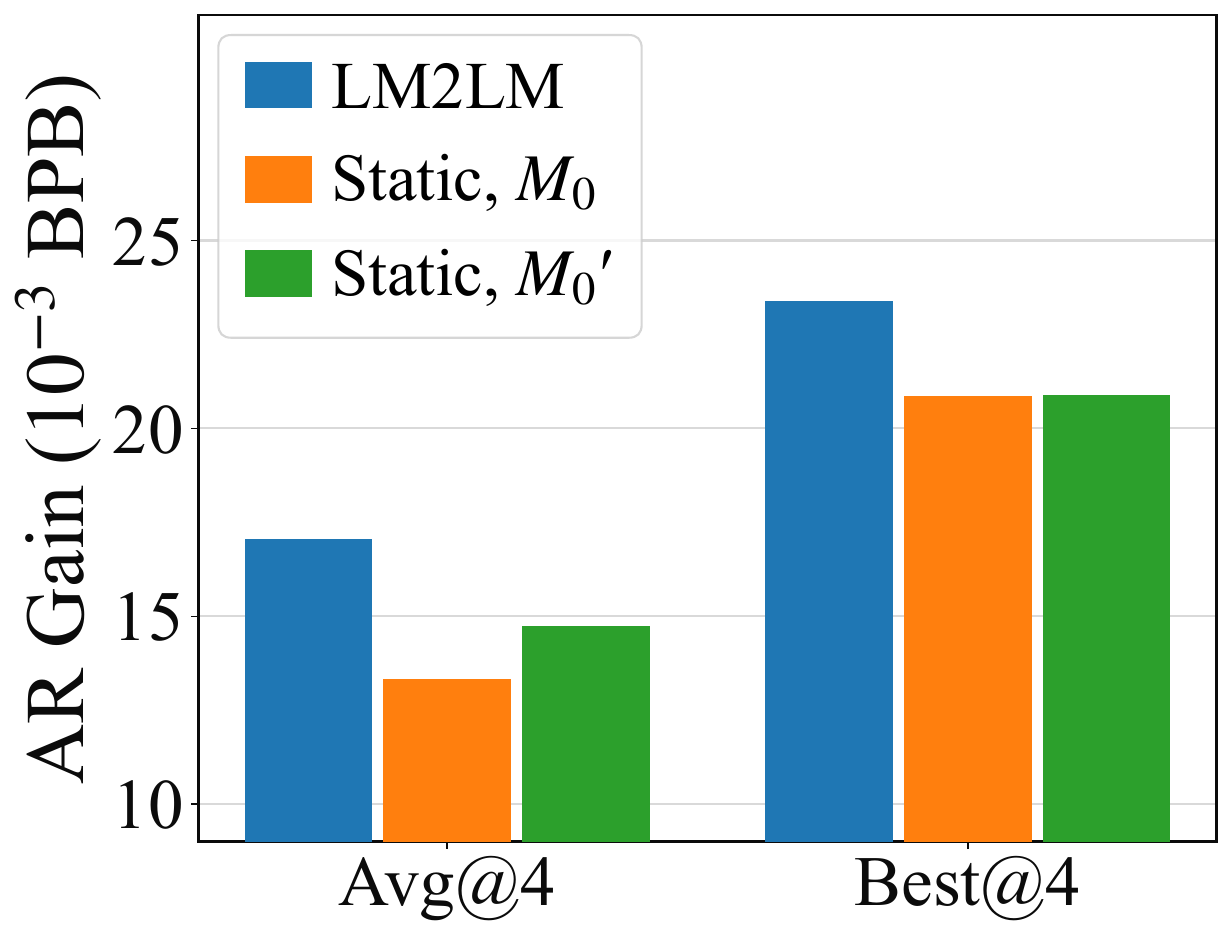}\caption{vs.\ static records.}\label{fig:ablation-static}\end{subfigure}\hspace{0.02\colfigwidth}
\begin{subfigure}[t]{0.49\colfigwidth}\centering\includegraphics[width=\linewidth]{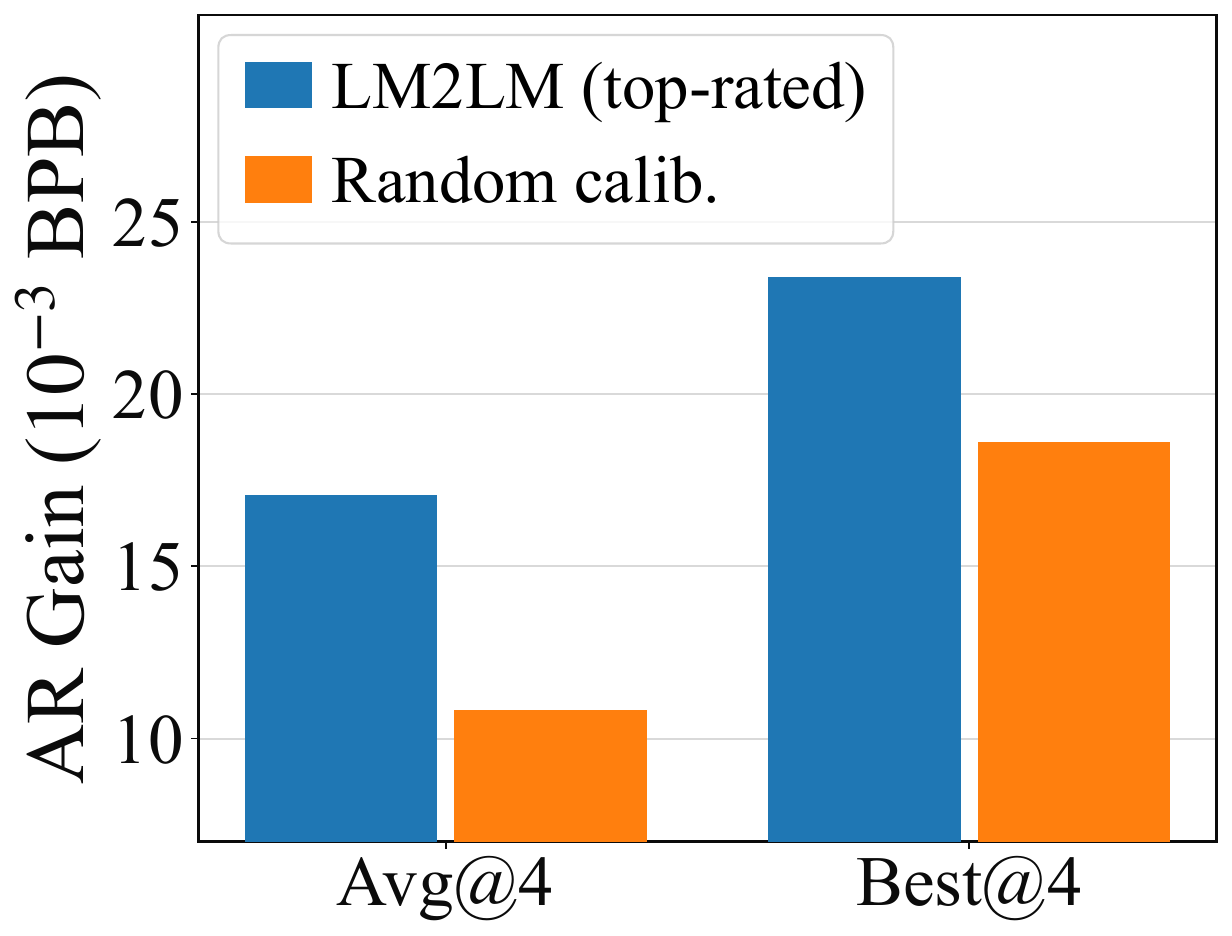}\caption{vs.\ random calibration.}\label{fig:ablation-random}\end{subfigure}
\caption{\textbf{Ablation of on-policy calibration on held-out AR research states at step 96.}}
\label{fig:ablation}
\vspace{-0.5em}
\end{figure}

\begin{figure*}[t]
\centering
\begin{subfigure}[t]{0.33\textwidth}\centering\includegraphics[width=\linewidth]{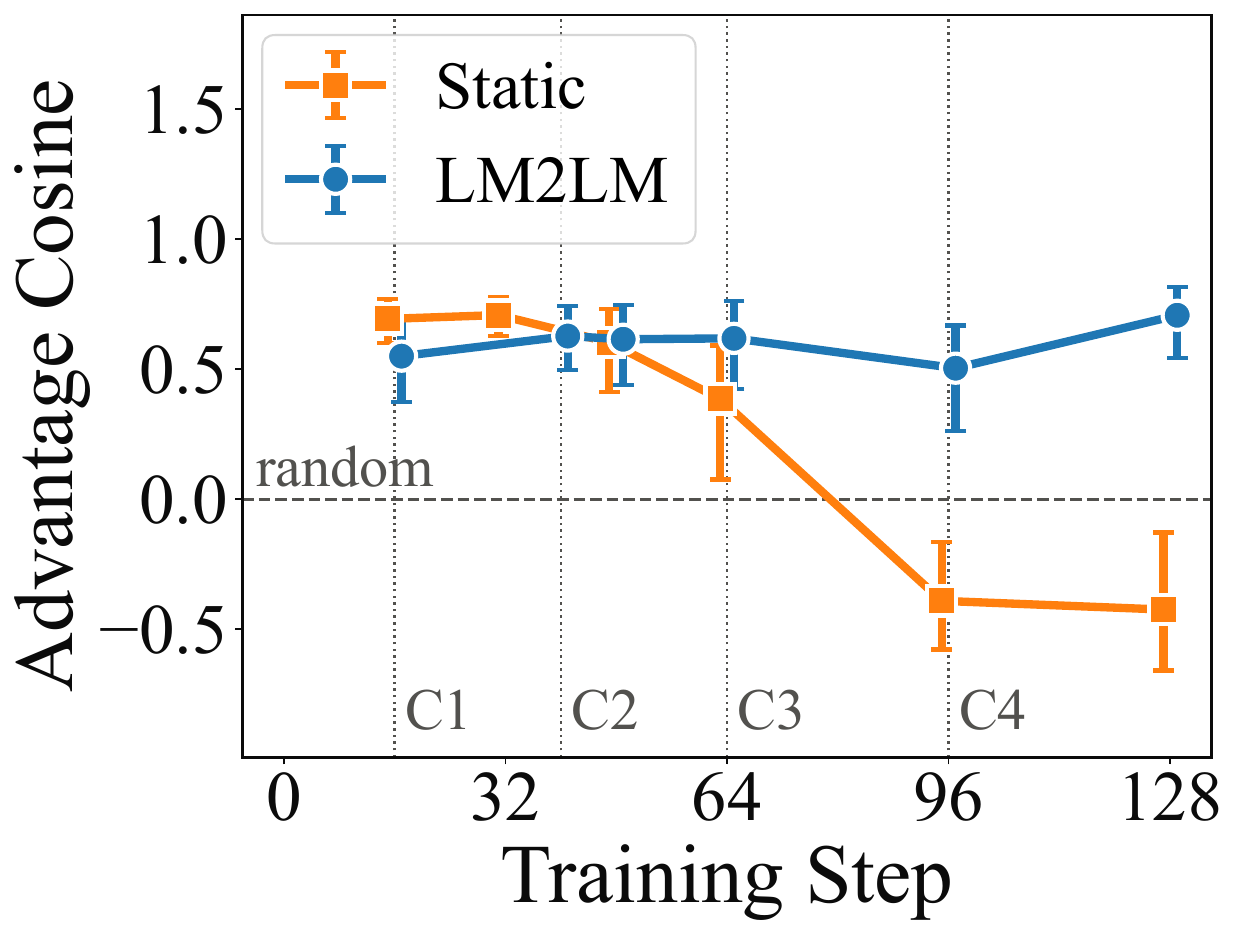}\caption{Advantage alignment.}\label{fig:adv-cos}\end{subfigure}\hfill
\begin{subfigure}[t]{0.33\textwidth}\centering\includegraphics[width=\linewidth]{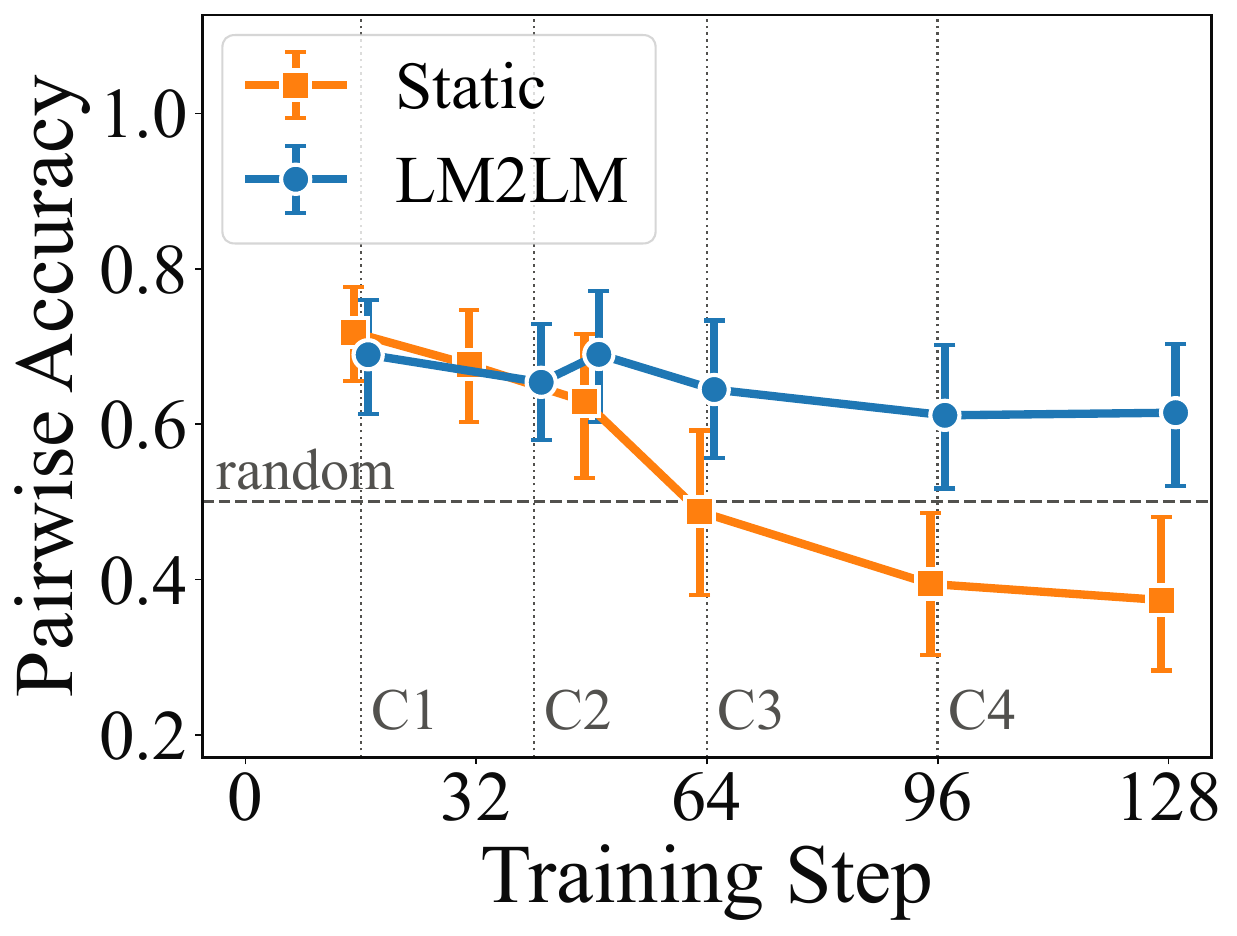}\caption{Pairwise accuracy.}\label{fig:adv-rank}\end{subfigure}\hfill
\begin{subfigure}[t]{0.33\textwidth}\centering\includegraphics[width=\linewidth]{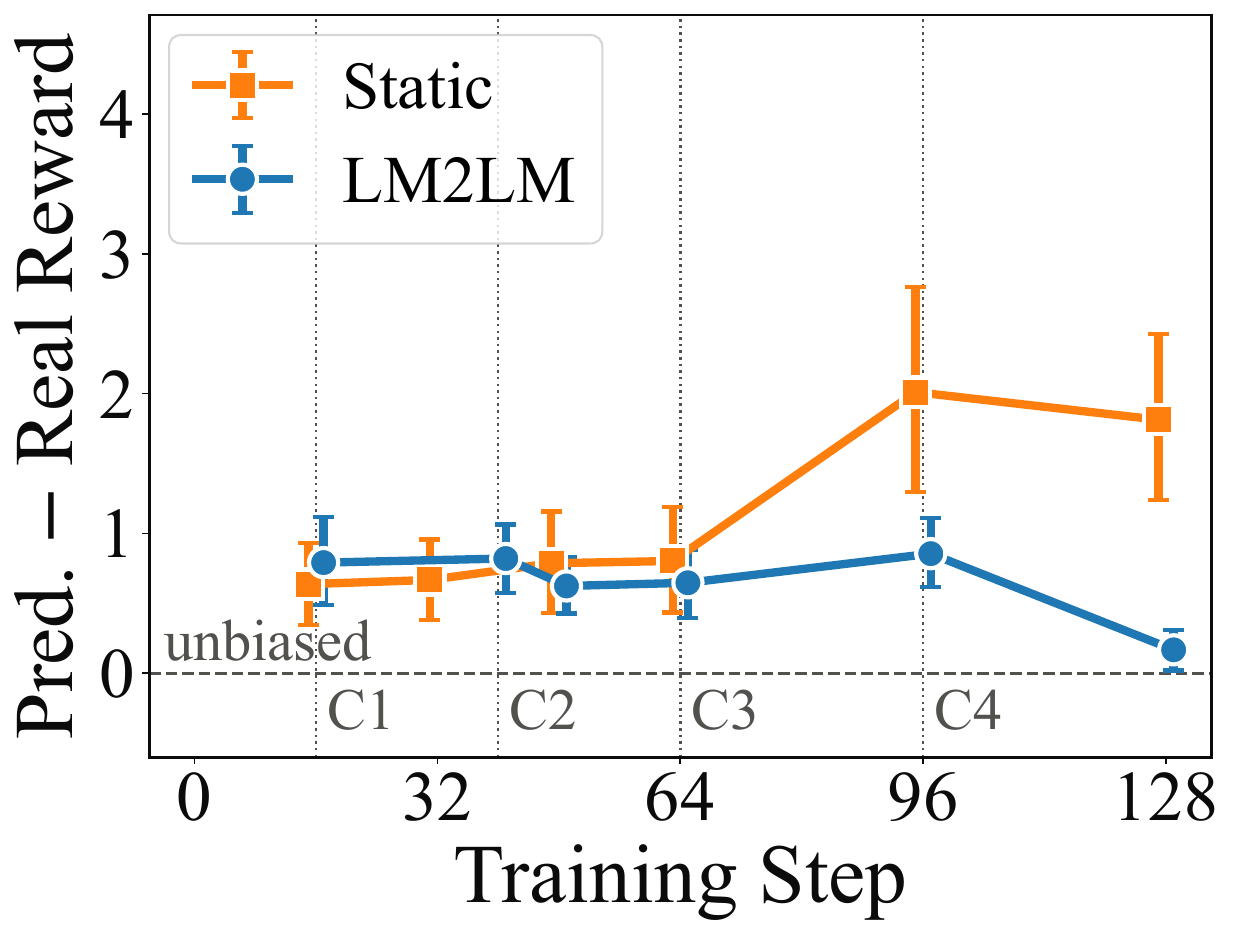}\caption{Predicted-real reward gap.}\label{fig:adv-gap}\end{subfigure}
\caption{\textbf{RWM feedback reliability on held-out AR research states}.}
\label{fig:adv}
\end{figure*}

\subsection{Generalization to Unseen Model Families}\label{ssec:RQ3}
We further evaluate whether \method generalizes to an unseen model family. Specifically, we hold out the entire Qwen3 family and train the proposer on the remaining families, excluding all Qwen3 records from training and RWM grounding.
(i) \emph{\method generalizes to unseen model families.}
Without Qwen3 records, \method achieves Avg@4 and Best@4 of 53.72 and 86.69, substantially outperforming the strongest baseline (21.63 and 25.27 for GPT-6-sol-ICL), demonstrating its ability to generate effective proposals for an unseen model family.
(ii) \emph{Out-of-family in-context records provide limited benefits.}
Base-ICL performs similarly to Base (2.97 vs.\ 3.53 Avg@4), while GPT-6-sol-ICL only slightly improves over GPT-6-sol (21.63 vs.\ 20.19).
This suggests that directly using out-of-family records offers limited benefits, highlighting the advantage of \method's RWM-based RL.

\subsection{Ablation Studies}\label{ssec:ablation}
To understand the contributions of on-policy calibration and the impact of its proposal selection strategy, we conduct ablation studies by comparing \method with variants that differ only in how the RWM records are constructed, with all other settings following Sec.~\ref{ssec:exp-env}.

\noindent\textbf{Impact of On-Policy Calibration.}
We compare \method with two static variants whose records stay fixed: $M_0$ (100 initial records) and $M_0'$ ($M_0$ plus 100 random records, matching \method's 200 real runs).
As shown in Fig.~\ref{fig:ablation}(a), \method achieves higher AR Avg@4 and Best@4 than both variants, and outperforms $M_0$ and $M_0'$ in Avg@4 on 74.2\% and 70.0\% of the held-out AR states, respectively.
With the same budget of 200 real runs, calibration outperforms adding random records (17.06 vs.\ 14.76 Avg@4), highlighting the importance of proposal selection for RWM grounding.

\noindent\textbf{Impact of Calibration Strategy.}
We further assess the impact of top-rated proposal selection in Eq.~\eqref{eq:select} by comparing \method with a variant that uniformly samples valid proposals whose outcomes are not yet in the records, while keeping the same initial records and three calibration rounds (200 real runs). As shown in Fig.~\ref{fig:ablation}(b), random calibration lowers AR Avg@4 and Best@4 from 17.06 and 23.40 to 10.84 and 18.60. 
We attribute this to top-rated selection correcting overestimated changes that RL is about to reinforce, whereas random selection can miss them or add misleading records.
Case studies are in Appendix~\ref{app:ablation}.

\subsection{In-depth Dissection of \method}\label{ssec:RQ2}

We further analyze why RWM feedback can guide RL and how calibration keeps it reliable. For this analysis, we extend training from 96 to 128 steps with a fourth calibration of \method (236 real runs) and compare it with the static variant whose RWM keeps $M_0$ throughout; each variant is scored by the RWM it is trained with.

\noindent\textbf{Why Can RWM Feedback Guide RL?}
We examine the alignment between RWM-predicted and real-execution feedback by measuring the cosine similarity of group-relative advantages and the within-state pairwise ranking accuracy over four executed proposals per held-out AR research state.
As shown in Figs.~\ref{fig:adv}(a) and (b), \method maintains advantage cosine of 0.50--0.71 and pairwise accuracy of 0.61--0.69, above chance levels of 0 and 0.5.
Since DAPO optimizes group-relative advantages, this indicates that the RWM provides informative relative feedback for policy optimization. 

\noindent\textbf{How Does On-Policy Calibration Help?}
To understand the gains from calibration in Sec.~\ref{ssec:ablation}, we examine how the feedback of the static variant evolves as RL changes the proposer's outputs (Fig.~\ref{fig:adv}). We observe that
(i) \emph{Calibration maintains reliable relative feedback.}
As shown in Fig.~\ref{fig:adv}(a), the static RWM's advantage cosine declines from 0.69 at step 16 to $-0.42$ at step 128, so its predicted advantages become negatively aligned with real ones, whereas \method keeps it positive.
Likewise, its pairwise accuracy drops from 0.72 to 0.37, below chance (0.5), while \method maintains 0.61 (Fig.~\ref{fig:adv}(b)).
We further observe that scoring the same \method proposals with $M_0$ instead of calibrated records lowers pairwise accuracy from 0.61 to 0.46 at step 96 (Appendix~\ref{app:same-proposals}).
These results indicate that on-policy calibration keeps the predicted advantages that DAPO optimizes aligned with real ones.
(ii) \emph{Calibration limits reward overestimation.} 
As shown in Fig.~\ref{fig:adv}(c), the predicted--real reward gap of the static RWM rises sharply after step 64, reaching about 2.0 at step 96, whereas \method's gap stays below 1 and decreases from 0.85 at step 96 to 0.17 at step 128 after the fourth calibration.
Thus, calibration also keeps predicted rewards close to real ones, mitigating reward overoptimization~\cite{gao2023scaling}. For example, our case studies in Appendix~\ref{app:feedback-cases} show that the initial RWM overestimates a NanoChat optimizer change (predicted $+0.7$ to $+1.25$ vs.\ real $-1.46$), so the static variant raises its share of NanoChat training proposals from 30\% to 79\% by step 64.
In contrast, the first calibration of \method adds its real outcomes to the records, and the share falls to 8\%.
More metrics and case studies are in Appendix~\ref{app:analysis}.

\section{Conclusion}\label{sec:conclusion}
We presented \method, which trains LLMs with RL to generate effective proposals for LM development, using a frozen LLM grounded in real experimental records as an RWM and keeping it reliable through on-policy calibration.
Extensive experiments across autoregressive and diffusion LM environments show that \method enables Qwen3.8-27B to outperform frontier models such as GPT-6-sol and achieve higher gains than baselines leveraging past experimental records at lower data-generation cost. Our analysis shows that RWM-based feedback provides informative rankings of proposal gains, while on-policy calibration mitigates reward overestimation and preserves ranking quality as training progresses. 
These findings highlight RWM-guided RL as a promising paradigm for learning research judgment from experimental feedback, offering a potential path toward RSI.

\section{Limitations}
Our study focuses on proposal-level improvements in LM development rather than end-to-end research campaigns.
Although the RWM provides informative rankings and substantially reduces the need for real execution during RL, its predictions remain imperfect, and on-policy calibration still requires a limited number of additional model-training runs.
Future work could extend our framework to broader research environments, including LM post-training, and evaluate its effectiveness in longer, end-to-end research campaigns.

\section{Ethics Statement}
This work studies the use of LLMs to automate research on language-model development.
Such systems may help reduce the cost of experimentation and accelerate scientific progress, but they also introduce risks, including unreliable experimental conclusions, increased computational resource consumption, and the potential misuse of automated model-improvement capabilities.
Our framework evaluates proposals in controlled training environments and uses real experimental outcomes to calibrate surrogate feedback.
Nevertheless, predicted rewards and experimental gains do not guarantee the safety or broader usefulness of proposed model changes.
We therefore emphasize the importance of human oversight, independent evaluation, and appropriate safeguards when applying automated research systems to more capable models or higher-stakes settings.




\clearpage
\bibliographystyle{assets/plainnat}
\bibliography{rsi}

\clearpage
\beginappendix

\section{More Details of Related Works}
\label{app:related}
\subsection{World Models for LLM Agents.}
\label{app:world_model}
A world model predicts the consequences of an agent's actions, enabling planning or learning without interacting with the real environment~\cite{sutton1991dyna,ha2018world,hafner2023mastering}. Recent work uses LLMs as world models for reasoning~\cite{hao2023reasoning}, web navigation~\cite{chae2025web,gu2024your}, text-based simulation~\cite{wang2024can}, and agent training in synthetic environments~\cite{chen2026scaling}.
In AutoResearch, concurrent work WMRL~\cite{yang2026scaling} uses a frozen LLM as the world model to predict execution outcomes as RL rewards and rescales them with a monotone mapping fitted on real executions at every training step.
In contrast, \method uses a frozen LLM as the world model that learns research knowledge from real experimental records in context~\cite{wang2026language}, and improves its predictions on the current policy's proposals by adding the outcomes of a few top-rated ones to these records.

\subsection{Reinforcement Learning for LLMs}
\label{app:rl-llm}
Reinforcement learning (RL)~\cite{shao2024deepseekmath,liu2025understanding,lin2025comprehensive,yu2026dapo,lin2026far,wu2026mmkgr1} has become an effective approach to improving LLM capabilities, particularly through reinforcement learning with verifiable rewards (RLVR)~\cite{lambert2025tulu}.
Methods such as GRPO~\cite{shao2024deepseekmath} and DAPO~\cite{yu2026dapo} optimize LLM policies using group-relative feedback without a learned value function.
While RLVR typically relies on inexpensive verification, such as checking mathematical answers or executing code, LM development requires costly model-training experiments to measure proposal quality.
Recent work applies RL to automated AI research using execution feedback~\cite{yang2026frontis,si2026towards}, but obtaining rewards through real execution at every training step limits scalability.
In contrast, \method uses RWM-predicted experimental gains as RL rewards, while periodically calibrating the RWM with selected real executions to maintain feedback reliability.

\subsection{Self-Evolving LLMs and Agents}
\label{app:self-evolve}
Self-evolving methods improve a model or agent with data or feedback it generates itself.
One line trains LLMs on self-generated solutions that are filtered or rewarded by verifiers or by the model itself~\cite{zelikman2022star,gulcehre2023reinforced,yuan2024self}.
Another line lets agents modify their own code, prompts, or workflows and keeps the changes that improve benchmark performance~\cite{lin2026position,yin2025godel,robeyns2025self,lin2026harness,lin2026memma}.
These methods rely on feedback that is cheap to obtain, such as answer checking, code execution, or benchmark scores.
In LM development, measuring a single proposal requires a full model-training run.
\method addresses this setting by learning from RWM-predicted gains and calibrating the RWM with a small number of real executions, and it targets the ability to improve LM development, which recursive self-improvement ultimately requires.

\section{Methodology Details of \method}\label{app:method}\label{app:impl}
\subsection{Reward Design}\label{app:impl-reward}
\noindent\textbf{Reward Scale.}
The task-specific reward scale is $\tau_e=\max(\sigma_e,\,1.4826\,\mathrm{MAD}_e)$, where $\sigma_e$ is the standard deviation of three reference runs and $\mathrm{MAD}_e$ is the median absolute deviation of the RWM-predicted gains for task $e$ from previous rollouts.
It is recomputed at the start of each training segment and held fixed within the segment; for the first segment, $\mathrm{MAD}_e$ is estimated from RWM predictions on proposals of a preliminary run.

\noindent\textbf{Validity, Repetition, and Bonuses.}
A proposal is valid if it can be translated into an executable modification of the task's reference configuration, which is checked without running it.
A proposal that changes nothing, i.e., all of whose modifications are inert, receives $-3$ instead of $-c$.
Repetition is decided on the resolved implementation: a proposal repeats the reference configuration if it resolves to it, the history if it matches a proposal in $h$, and an earlier proposal of its group if their implementations coincide.
A proposal that repeats an earlier proposal of its own group keeps at most zero reward, and if no proposal of a group is valid and novel, every proposal of the group receives $-c$.
In the rollouts whose proposals are selected for calibration, a valid proposal with a positive predicted gain also receives a novelty bonus $0.2\,\sigma_G/\sqrt{1+N_{\mathrm{prev}}}$ in addition to the diversity bonus $b_i$, where $N_{\mathrm{prev}}$ is the number of earlier occurrences of the same change on the task.

\noindent\textbf{Hyperparameters.}
We use a repetition margin $\delta=1$ and a diversity weight $\lambda=0.3$.

\subsection{Synthetic Training-State Expansion}\label{app:impl-refresh}
Training states are initially constructed as described in Appendix~\ref{app:data-construction}.
After each rollout, i.e., every 8 training steps, we add the proposals generated in that rollout to the history pool, with their gains predicted by the current RWM as synthetic outcomes, and resample histories from the combined pool of real and predicted records to construct the training states of the next rollout.

\subsection{Algorithm of \method}\label{app:alg}
Algorithm~\ref{alg:method} summarizes the full training procedure of \method.

\begin{algorithm}[t]
\caption{\method}\label{alg:method}
\renewcommand{\algorithmicrequire}{\textbf{Input:}}
\renewcommand{\algorithmicensure}{\textbf{Output:}}
\begin{algorithmic}[1]
\REQUIRE base proposer $\pi_{\theta_0}$, RWM $W$, builder $\mathcal B$, training tasks $\mathcal E_{\mathrm{train}}$, number of initial records $N_0$, number of calibrations $U$, calibration batch sizes $\{B_u\}_{u=0}^{U-1}$
\ENSURE trained proposer $\pi_\theta$
\STATE Execute $N_0$ proposals on $\mathcal E_{\mathrm{train}}$ to obtain the initial records $M_0$
\STATE Build the training-state distribution $\mu_{\mathrm{train}}$ from $M_0$
\FOR{$u=0,\dots,U$}
  \FOR{each RL step of training segment $u$}
    \STATE Sample states $s\sim\mu_{\mathrm{train}}$ and $G$ proposals per state from $\pi_\theta$
    \STATE Compute rewards $R_i$ with the RWM $W(\cdot\,;M_u)$ \COMMENT{Eq.~(\ref{eq:rl-reward})}
    \STATE Update $\theta$ with DAPO \COMMENT{Eq.~(\ref{eq:dapo})}
    \STATE Periodically refresh $\mu_{\mathrm{train}}$ with earlier proposals and their predicted gains
  \ENDFOR
  \IF{$u<U$}
    \STATE Select the top-rated batch $\mathcal Q_u$ of $B_u$ proposals of the current policy \COMMENT{Eq.~(\ref{eq:select})}
    \STATE Implement and execute $\mathcal Q_u$ with $\mathcal B$ to measure their real gains
    \STATE $M_{u+1}\gets M_u\cup\{(e,a,g(s,a)) : (s,a)\in\mathcal Q_u\}$ \COMMENT{Eq.~(\ref{eq:update})}
  \ENDIF
\ENDFOR
\RETURN $\pi_\theta$
\end{algorithmic}
\end{algorithm}

\begin{table*}[t]
\centering
\small
\setlength{\tabcolsep}{5pt}

\caption{\textbf{Training and held-out tasks, evaluation states, and experimental records by model family}. Parentheses indicate the number of states per held-out task; $M_0$ and Pool denote the initial RWM records and the 300-record baseline dataset, respectively.}
\label{tab:data}
\resizebox{\textwidth}{!}{
\begin{tabular}{l p{7.2cm} p{5.0cm} c cc}
\toprule
& & & \textbf{Held-out} & \multicolumn{2}{c}{\textbf{Records}} \\
\cmidrule(lr){5-6}
\textbf{Family} & \textbf{Training tasks} & \textbf{Held-out tasks (states)} & \textbf{states} & $M_0$ & \textbf{Pool} \\
\midrule
\multicolumn{6}{@{}l}{\emph{AR families}} \\
OLMo-3    & \texttt{100m-4c}, \texttt{100m-8c}, \texttt{190m-1c}, \texttt{190m-2c}, \texttt{370m-1c}, \texttt{370m-2c}, \texttt{760m-1c} & \texttt{100m-1c} (17), \texttt{100m-2c} (8), \texttt{190m-4c} (4) & 29 & 16 & 47 \\
Qwen3     & \texttt{q06b-1b}, \texttt{q06b-2b}, \texttt{q17b-1b}, \texttt{q17b-2b}, \texttt{q152m-8b}, \texttt{q267m-1b}, \texttt{q267m-2b} & \texttt{q152m-2b} (8), \texttt{q267m-4b} (4) & 12 & 25 & 62 \\
NanoChat  & \texttt{d8-8k-0p82b}, \texttt{d8-8k-3p28b}, \texttt{d12-8k-8p59b}, \texttt{d16-8k-4p58b}, \texttt{d20-8k-8p49b} & \texttt{d12-8k-2p15b} (17) & 17 & 15 & 79 \\
Marin     & \texttt{153m-2p9b}, \texttt{153m-8b}, \texttt{267m-2p9b}, \texttt{267m-8b}, \texttt{486m-2p9b} & \texttt{486m-8b} (2) & 2  & 11 & 30 \\
\midrule
\multicolumn{6}{@{}l}{\emph{Diffusion families}} \\
E2D2      & \texttt{34m-6e18}, \texttt{113m-3e18}, \texttt{113m-6e18}, \texttt{231m-6e18} & \texttt{66m-1e18} (6) & 6  & 11 & 31 \\
BD3LM     & \texttt{34m-6e18}, \texttt{113m-3e18}, \texttt{113m-1e19}, \texttt{231m-6e18} & \texttt{113m-6e18} (6) & 6  & 7  & 19 \\
MDM       & \texttt{34m-6e18}, \texttt{85m-6e18}, \texttt{85m-1e19}, \texttt{231m-6e18} & -- & 0  & 6  & 14 \\
E2D2-SW & \texttt{34m-6e18}, \texttt{113m-3e18}, \texttt{113m-1e19}, \texttt{206m-6e18} & \texttt{113m-6e18} (6) & 6  & 9  & 18 \\
\midrule
Total     & 40 training tasks & 10 evaluated held-out tasks & 78 & 100 & 300 \\
\bottomrule
\end{tabular}}
\end{table*}

\begin{table*}[t]
\centering
\small
\setlength{\tabcolsep}{6pt}

\caption{\textbf{Held-out tasks, model sizes, training budgets, and reference statistics}. Parameters are total model parameters; budgets are training tokens (AR) or FLOPs with corresponding tokens (diffusion). $b_e$ and $\sigma_e$ are the mean and standard deviation of reference validation loss over three runs, in BPB (AR) and nats per token (diffusion).}

\label{tab:heldout}
\resizebox{0.7\textwidth}{!}{
\begin{tabular}{ll cccc}
\toprule
\textbf{Family} & \textbf{Task} & \textbf{Parameters} & \textbf{Budget} & $b_e$ & $\sigma_e$ \\
\midrule
OLMo-3    & \texttt{100m-1c}      & 153M & 2.0B tokens  & 1.3750 & 0.0011 \\
OLMo-3    & \texttt{100m-2c}      & 153M & 4.0B tokens  & 1.2939 & 0.0056 \\
OLMo-3    & \texttt{190m-4c}      & 267M & 15.2B tokens & 1.1275 & 0.0017 \\
Qwen3     & \texttt{q152m-2b}     & 153M & 2.0B tokens  & 1.3641 & 0.0025 \\
Qwen3     & \texttt{q267m-4b}     & 267M & 4.0B tokens  & 1.1683 & 0.0042 \\
NanoChat  & \texttt{d12-8k-2p15b} & 286M & 2.15B tokens & 0.8135 & 0.0007 \\
Marin     & \texttt{486m-8b}      & 486M & 8.0B tokens  & 0.9959 & 0.0008 \\
\midrule
E2D2      & \texttt{66m-1e18}     & 115M & $1\times10^{18}$ FLOPs (1.55B tokens) & 3.5508 & 0.0037 \\
BD3LM     & \texttt{113m-6e18}    & 162M & $6\times10^{18}$ FLOPs (3.46B tokens) & 3.1218 & 0.0005 \\
E2D2-SW & \texttt{113m-6e18}    & 196M & $6\times10^{18}$ FLOPs (2.44B tokens) & 3.2846 & 0.0026 \\
\bottomrule
\end{tabular}}
\end{table*}

\section{Environments and Data}\label{app:env-data}

\subsection{Model Families}\label{app:env}
Each family shares an architecture, training-data distribution, and evaluation metric across its tasks.

\noindent\textbf{AR Families.}
\begin{itemize}[leftmargin=*]
    \item \emph{OLMo-3}~\cite{olmo2025olmo}: OLMo-3 small models trained on samples of Dolma~3 and evaluated by mean bits per byte (BPB) across ten downstream tasks (ARC-Challenge, ARC-Easy, HellaSwag, four MMLU subsets, HumanEval, MBPP, and MATH-500).
    \item \emph{Qwen3}~\cite{yang2025qwen3}: The official Qwen3 architecture and tokenizer. Training data (Dolma~3) and evaluation (ten-task BPB) are shared with OLMo-3.
    \item \emph{NanoChat}~\cite{nanochat}: depth-parameterized transformers trained on ClimbMix and evaluated by BPB on a held-out ClimbMix validation shard.
    \item \emph{Marin}~\cite{hall2025marin}: Marin architectures with the same training data and evaluation as OLMo-3.
\end{itemize}

\noindent\textbf{Diffusion Families.}
All four families are trained on SlimPajama with a 2,048-token context and evaluated by the negative log-likelihood of true tokens accumulated along the model's greedy, most-confident-first unmasking order, measured in nats per token.
\begin{itemize}[leftmargin=*]
    \item \emph{E2D2}~\cite{arriola2026encoder}: an encoder-decoder masked diffusion model with a clean-token encoder and a cross-attention decoder for noisy tokens. Noise levels are sampled per block, yielding block-autoregressive modeling.
    \item \emph{E2D2-SW}~\cite{wang2026revisiting}: This is an E2D2 variant with a token-causal encoder and a decoder trained on packed sliding windows of blocks but evaluated on non-overlapping blocks.
    \item \emph{BD3LM}~\cite{arriola2025block}: a block diffusion model using a single network over concatenated clean and noisy sequences with block-structured attention.
    \item \emph{MDM}~\cite{nie2025scaling}: a full-sequence bidirectional masked diffusion model with one noise level per sequence; it appears only among the training tasks and has no held-out task.
\end{itemize}

\noindent\textbf{Tasks.}
A task specifies a model family, size, and training budget. Tasks within a family share the training data, tokenizer, evaluation protocol, and hardware. We construct tasks across these model families, following the naming conventions below:
\begin{itemize}[leftmargin=*]
    \item \emph{OLMo-3}: \texttt{100m-1c} denotes 100M non-embedding parameters trained with $C=1$ times the Chinchilla-optimal budget~\cite{hoffmann2022training}, or $20C$ tokens per parameter (2B tokens).
    \item \emph{Qwen3}: Two model lines are kept separate. \texttt{q152m-2b} has 152.5M parameters, 2B training tokens, and an 8,192-token context. \texttt{q06b-1b} uses the official Qwen3-0.6B configuration with 1B training tokens and a 2,048-token context, appearing only in training tasks.
    \item \emph{NanoChat}: \texttt{d12-8k-2p15b} denotes a depth-12 model (width 768) with an 8,192-token context trained on 2.15B tokens; \texttt{p} denotes the decimal point.
    \item \emph{Marin}: \texttt{486m-8b} denotes 485.9M parameters trained on 8B tokens; \texttt{2p9b} denotes 2.9B tokens.
    \item \emph{Diffusion families}: \texttt{113m-6e18} denotes 113M non-embedding parameters trained with $6\times10^{18}$ FLOPs, from which the token budget is derived.
\end{itemize}

Each task uses its family's default training recipe as the reference configuration. The reference loss $b_e$ and seed noise $\sigma_e$ are the mean and standard deviation of validation loss over three runs.
Table~\ref{tab:data} summarizes training tasks, held-out tasks and states, and available executed records by family. Table~\ref{tab:heldout} details the held-out tasks, which differ from training tasks in model size or training budget.

\begin{table*}[t]
\centering
\small

\caption{\textbf{Distribution of pre-collected experimental records across mechanism directions for AR and diffusion families}. Each proposal is counted once under its primary direction. Other includes head shape and context-window curriculum.}

\label{tab:directions}
\resizebox{\textwidth}{!}{\begin{tabular}{l l cc}
\toprule
\textbf{Direction} & \textbf{Example changes} & \textbf{AR} & \textbf{Diffusion} \\
\midrule
Attention reach & RoPE fraction and base, attention windows, token-mixer layout & 42 & -- \\
Decay shape & learning-rate decay tail, warmup, weight averaging, $\beta_2$ & 34 & -- \\
Update geometry & Muon iterations and scaling, $\beta_2$, weight decay, gradient clipping & 22 & 9 \\
Capacity allocation & per-layer FFN widths, attention--FFN parameter split & 29 & -- \\
KV / state budget & placement and slots of context-memory layers, KV heads & 23 & -- \\
Loss geometry & confidence-weighted loss, label smoothing, position-scheduled loss & 23 & -- \\
Prediction horizon & multi-token prediction heads and auxiliary weights & 21 & -- \\
Residual / norm placement & normalizer form, normalization sites, QK normalization & 21 & -- \\
LR schedule & learning rate, schedule shape, warmup, minimum learning rate & -- & 27 \\
Mask process & mask $\epsilon$, mask schedule, low-variance loss & -- & 15 \\
Backbone form & query groups, rotary fraction, norm and MLP class & -- & 14 \\
Block structure & multi-token window, decoder depth, encoder causality & -- & 13 \\
Auxiliary objective & AR auxiliary loss, encoder causality, loss-weight schedule & -- & 4 \\
Other & head shape, context-window curriculum & 3 & -- \\
\midrule
Total & & 218 & 82 \\
\bottomrule
\end{tabular}}
\end{table*}

\begin{table*}[t]
\centering\small
\caption{\textbf{A representative mechanism card from the library}.}
\label{tab:app-card}
\begin{promptbox}[Card M-DEC-01: learning-rate decay tail]
\textbf{Direction.} Decay shape. \textbf{Mechanism.} Learning-rate decay tail: schedule family, decay fraction, and final learning-rate floor.

\textbf{Hypothesis.} Moving the decay tail from the reference toward cosine or a lower final floor helps where the reference keeps a non-zero floor, with a gain that grows with the token budget; it is neutral or harmful where the reference already decays to about zero (Marin, linear to zero).

\textbf{Prediction.} Positive. Stronger on tasks with many tokens per parameter (e.g., OLMo-3 \texttt{100m-8c}, \texttt{190m-2c}, \texttt{370m-2c}); weaker on tasks with few tokens per parameter (e.g., OLMo-3 \texttt{760m-1c}) and on Marin.

\textbf{Evidence.} In earlier single-mechanism runs on training tasks, the change improved all 15 OLMo-3 runs and 3 of 4 Qwen3 runs, but only 1 of 8 Marin runs.

\textbf{Levels.} OLMo-3: cosine schedules with final floors 0, 0.01, and 0.1, and a warmup-stable-decay schedule (4 levels). NanoChat: cosine to floors 0, 0.01, and 0.1, and linear to 0 (4). Qwen3: 7 schedule settings. Marin: 6 schedule settings.
\end{promptbox}
\end{table*}

\subsection{Data Construction}
\label{app:data-construction}\label{app:cards}

\noindent\textbf{Mechanism Library.}
We curate a library of training-recipe mechanisms covering model architecture, optimization, and training objectives.
Each reviewed card specifies a hypothesis, expected effects, supporting evidence, and family-specific configurations (levels) with validity constraints.
The AR library contains 40 accepted cards; Table~\ref{tab:app-card} shows an example.

The library also defines the action space shared by all proposers, specifying modifiable recipe settings and their valid values.
Proposals may combine any number of mechanisms, choose numerical values within valid ranges, and propose new variants for allowed categorical settings, which the builder implements.
Model size, training budget, data, tokenizer, evaluation, seeds, and hardware remain fixed.

\noindent\textbf{Experimental Records.}
We construct 300 proposal--outcome records across 40 training tasks.
To ensure diversity and balanced coverage of mechanism directions, we sample proposals by instantiating one or two mechanism-card levels per proposal.
Of the 300 executed proposals, 226 involve a single mechanism direction and 74 combine two (Table~\ref{tab:directions}).
We select 100 records as the initial RWM records $M_0$.

\noindent\textbf{Training and Baseline Data.}
Each research state $s=(e,h)$ consists of a task $e$ and a history $h$ of proposals and their outcomes.
Initial RL histories contain 0--8 task-matched real records sampled from $M_0$, and are expanded with RWM-predicted outcomes during RL (Appendix~\ref{app:impl-refresh}).
ICL baselines use up to 300 executed records as in-context examples.
SFT uses 300 $(s,a)$ pairs constructed from these records, while RFT-Real retains the 114 pairs with positive measured gains.

\noindent\textbf{Held-out States.}
We construct 60 AR and 18 diffusion research states from 10 held-out tasks.
All states contain histories of real experimental outcomes and are fixed across methods.
No held-out experimental records are used for proposer training, RWM grounding, or calibration.

\begin{table*}[t]
\centering
\small
\setlength{\tabcolsep}{6pt}
\caption{\textbf{Per-family results on held-out AR tasks using real execution}.
Avg@4 and Best@4 are reported in $10^{-3}$ BPB.
Win rate compares the first model against the second, with values above 0.5 favoring the first.
Best results in each family are in \textbf{bold}.}
\label{tab:per-family-ar}
\resizebox{0.7\textwidth}{!}{
\begin{tabular}{l|cc|cc}
\toprule
\multirow{2}{*}{\textbf{Model}}
& \multicolumn{2}{c|}{\textbf{Experimental Gain} $\uparrow$}
& \multicolumn{2}{c}{\textbf{Pairwise Win Rate} $\uparrow$} \\
\cmidrule(lr){2-3}\cmidrule(lr){4-5}
& \textbf{Avg@4} & \textbf{Best@4}
& \textbf{Row vs.\ Base} & \textbf{\method\ vs.\ Row} \\
\midrule
\multicolumn{5}{@{}l}{\emph{OLMo-3}} \\
Base (Qwen3.8-27B) & $-10.72$ & $10.48$ & -- & $0.862$ \\
Base-ICL & $-3.58$ & $10.97$ & $0.724$ & $0.759$ \\
GPT-6-sol & $5.54$ & $8.26$ & $0.828$ & $0.586$ \\
GPT-6-sol-ICL & $-1.16$ & $6.21$ & $0.655$ & $0.655$ \\
SFT & $-22.03$ & $1.70$ & $0.483$ & $0.931$ \\
RFT-Real & $-16.49$ & $5.55$ & $0.448$ & $0.931$ \\
\method & $\mathbf{7.39}$ & $\mathbf{17.02}$ & $0.862$ & -- \\
\midrule
\multicolumn{5}{@{}l}{\emph{Qwen3}} \\
Base (Qwen3.8-27B) & $3.53$ & $31.39$ & -- & $1.000$ \\
Base-ICL & $31.27$ & $52.40$ & $0.917$ & $1.000$ \\
GPT-6-sol & $20.19$ & $22.75$ & $0.667$ & $0.917$ \\
GPT-6-sol-ICL & $29.76$ & $36.61$ & $0.750$ & $0.917$ \\
SFT & $4.68$ & $23.59$ & $0.583$ & $1.000$ \\
RFT-Real & $3.35$ & $32.37$ & $0.583$ & $1.000$ \\
\method & $\mathbf{67.58}$ & $\mathbf{74.26}$ & $1.000$ & --  \\
\midrule
\multicolumn{5}{@{}l}{\emph{NanoChat}} \\
Base (Qwen3.8-27B) & $-11.28$ & $-1.92$ & -- & $1.000$ \\
Base-ICL & $-2.24$ & $0.66$ & $0.941$ & $0.235$ \\
GPT-6-sol & $-1.40$ & $-0.73$ & $1.000$ & $0.294$ \\
GPT-6-sol-ICL & $\mathbf{0.74}$ & $\mathbf{1.23}$ & $1.000$ & $0.000$ \\
SFT & $-7.64$ & $0.10$ & $0.647$ & $0.882$ \\
RFT-Real & $-50.55$ & $-0.23$ & $0.412$ & $0.941$ \\
\method & $-1.89$ & $-1.07$ & $1.000$ & -- \\
\midrule
\multicolumn{5}{@{}l}{\emph{Marin}} \\
Base (Qwen3.8-27B) & $-8.12$ & $17.75$ & -- & $1.000$ \\
Base-ICL & $4.60$ & $17.28$ & $0.500$ & $1.000$ \\
GPT-6-sol & $5.85$ & $13.65$ & $0.500$ & $0.500$ \\
GPT-6-sol-ICL & $\mathbf{16.56}$ & $18.01$ & $1.000$ & $0.500$ \\
SFT & $-9.52$ & $-1.61$ & $0.500$ & $1.000$ \\
RFT-Real & $-39.77$ & $18.11$ & $0.000$ & $1.000$ \\
\method & $15.15$ & $\mathbf{18.61}$ & $1.000$ & -- \\
\bottomrule
\end{tabular}}
\end{table*}

\begin{table*}[t]
\centering
\small
\setlength{\tabcolsep}{6pt}

\caption{\textbf{Per-family results on held-out diffusion tasks using real execution.}
Avg@4 and Best@4 are reported in $10^{-2}$ nats per token.
Win rate compares the first model against the second, with values above 0.5 favoring the first.
Best results in each family are in \textbf{bold}.}

\label{tab:per-family-diffusion}
\resizebox{0.7\textwidth}{!}{
\begin{tabular}{l|cc|cc}
\toprule
\multirow{2}{*}{\textbf{Model}}
& \multicolumn{2}{c|}{\textbf{Experimental Gain} $\uparrow$}
& \multicolumn{2}{c}{\textbf{Pairwise Win Rate} $\uparrow$} \\
\cmidrule(lr){2-3}\cmidrule(lr){4-5}
& \textbf{Avg@4} & \textbf{Best@4}
& \textbf{Row vs.\ Base} & \textbf{\method\ vs.\ Row} \\
\midrule
\multicolumn{5}{@{}l}{\emph{E2D2}} \\
Base (Qwen3.8-27B) & $-5.37$ & $7.65$ & -- & $1.000$ \\
Base-ICL & $3.28$ & $6.88$ & $0.667$ & $0.667$ \\
GPT-6-sol & $3.23$ & $6.12$ & $0.667$ & $0.667$ \\
GPT-6-sol-ICL & $3.72$ & $4.98$ & $0.833$ & $0.833$ \\
SFT & $-4.95$ & $6.83$ & $0.500$ & $1.000$ \\
RFT-Real & $-6.64$ & $2.69$ & $0.500$ & $1.000$ \\
\method & $\mathbf{7.30}$ & $\mathbf{9.67}$ & $1.000$ & -- \\
\midrule
\multicolumn{5}{@{}l}{\emph{BD3LM}} \\
Base (Qwen3.8-27B) & $-3.36$ & $6.62$ & -- & $1.000$ \\
Base-ICL & $2.64$ & $6.23$ & $0.833$ & $0.833$ \\
GPT-6-sol & $4.27$ & $5.43$ & $1.000$ & $0.833$ \\
GPT-6-sol-ICL & $3.13$ & $3.17$ & $1.000$ & $1.000$ \\
SFT & $-0.33$ & $2.53$ & $0.833$ & $1.000$ \\
RFT-Real & $-1.34$ & $4.80$ & $0.500$ & $0.833$ \\
\method & $\mathbf{7.31}$ & $\mathbf{9.81}$ & $1.000$ & -- \\
\midrule
\multicolumn{5}{@{}l}{\emph{E2D2-SW}} \\
Base (Qwen3.8-27B) & $-3.82$ & $3.40$ & -- & $1.000$ \\
Base-ICL & $6.18$ & $\mathbf{12.38}$ & $1.000$ & $0.667$ \\
GPT-6-sol & $-1.55$ & $1.73$ & $0.833$ & $1.000$ \\
GPT-6-sol-ICL & $\mathbf{7.96}$ & $9.65$ & $1.000$ & $0.333$ \\
SFT & $-3.99$ & $1.82$ & $0.500$ & $1.000$ \\
RFT-Real & $-3.18$ & $2.62$ & $0.500$ & $0.833$ \\
\method & $6.70$ & $10.29$ & $1.000$ & -- \\
\bottomrule
\end{tabular}}
\end{table*}

\section{Evaluation Protocol}\label{app:eval-prot}
\subsection{Metrics}\label{app:metrics}
We evaluate all proposers on the same fixed set of 60 autoregressive (AR) and 18 diffusion held-out research states.
For each state, a proposer generates four proposals, which are implemented by the same builder and evaluated through real model-training experiments.
Invalid or unchanged proposals receive zero gain, while negative gains are retained.

\noindent\textbf{Validation Metrics.}
We measure model performance using:
\begin{itemize}[leftmargin=*]
    \item \textbf{BPB (Bits per Byte):} Average number of bits needed to encode each byte of text, enabling comparisons across tokenizers~\cite{gao2020pile}. We use it for AR models, evaluated on held-out text as specified per family (Appendix~\ref{app:env}).
    \item \textbf{NLL (Negative Log-Likelihood):} Negative log-probability assigned to validation sequences. For diffusion models, we compute the exact negative log-likelihood of the true tokens accumulated along the model's greedy, most-confident-first within-block unmasking order, with each revealed position set to the true token, reported in nats per token.
\end{itemize}
Lower values indicate better model performance.

\noindent\textbf{Evaluation Metrics.}
We evaluate proposal quality using:
\begin{itemize}[leftmargin=*]
    \item \textbf{Avg@4 / Best@4:} Mean / maximum real gain among four proposals per state, averaged equally across states.
    \item \textbf{Win rate:} Fraction of states where one method achieves higher Avg@4 than another, with ties counted as 0.5.
    \item \textbf{Neg.:} Fraction of proposals with negative real gains.
\end{itemize}
Experimental gains are reported in $10^{-3}$ BPB (AR) and $10^{-2}$ nats per token (diffusion).

\subsection{Cost Accounting}\label{app:cost-accounting}

\noindent\textbf{Hardware and Prices.}
All experiments, including baseline training and evaluation, RL training, experimental data collection, reference runs, and on-policy calibration, use NVIDIA B200 GPUs.
GPU usage is priced at \$14.208 per B200 GPU-hour\footnote{\url{https://aws.amazon.com/ec2/capacityblocks/pricing/}}.
For GPT-6-sol RWM calls, token prices\footnote{\url{https://developers.openai.com/api/docs/models/gpt-6-sol}} per million are \$2 (uncached input), \$0.20 (cached input), \$2.50 (cache write), and \$10 (output).

\noindent\textbf{Data-Generation Cost.}
We measure cost as the GPU expenses of real experimental executions plus the token cost of RWM predictions.
Real executions include those used to collect ICL records, fine-tuning data, initial RWM records, and on-policy calibration records, including failed runs.
We estimate RWM prediction costs using the corresponding model's token prices.
We exclude proposer and builder inference, proposer fine-tuning, and shared research-state construction.

\noindent\textbf{Data-Generation Time.}
We measure time as the sum of real-execution durations, assuming sequential execution, and RWM prediction time.

\section{Additional Implementation Details}
\label{sec:exp-impl-details}

\subsection{Models and Prompts.}
The proposer's backbone is Qwen3.8-27B.
The RWM, GPT-6-sol, is queried with a single API request per proposal at high reasoning effort, without tools.
The builder, GPT-6-astra, runs as a Codex agent~\cite{openai2025codex} (\texttt{codex exec}) at high reasoning effort. 
Tables~\ref{tab:prompt-rwm}, and~\ref{tab:prompt-builder} list the prompts of RWM, and builder, respectively.

\subsection{RL Training}\label{app:impl-rl-train}
We use DAPO with asymmetric clipping ($\epsilon_{\mathrm{low}}=0.2$, $\epsilon_{\mathrm{high}}=0.28$), a token-level policy-gradient loss, and no KL or entropy regularization.
For dynamic sampling, we sample 277 states and retain 256 per rollout; importance-sampling ratios are clipped at 2.0.
Proposals are sampled at temperature 1.0 with a maximum of 20{,}000 generated tokens.
Each rollout is followed by eight policy updates with a global batch size of 256.
We use Adam ($\beta_1=0.9$, $\beta_2=0.98$), weight decay 0.1, and an initial learning rate of $10^{-6}$, multiplied by 0.7 after each training segment.

\subsection{On-Policy Calibration}\label{app:impl-cal}
The candidates $\mathcal C_u$ in Eq.~(\ref{eq:select}) are the proposals sampled by the current proposer on training states whose outcomes are not yet in $M_u$; in practice, we spread the selected top-rated proposals across training tasks and states.
Each selected proposal is executed once on its training task, and the three calibrations add $B_0=31$, $B_1=38$, and $B_2=31$ records.
Calibration outcomes enter only the RWM records, not the histories of training states.

\subsection{SFT and RFT-Real}\label{app:impl-sft}
Each SFT pair uses a record's executed proposal as the target, and a state of the record's own task whose history contains 0--7 other records of that task from the 300 records, chosen to diversify histories across pairs.
RFT-Real keeps the 114 pairs whose measured gain is positive (76 AR and 38 diffusion).
We train separate LoRA adapters for AR and diffusion tasks with the configuration in Table~\ref{tab:sft-config}, and sample proposals with the same decoding settings as \method.

\section{Additional Main Results}\label{app:main-results}

\subsection{Per-Family Results}\label{app:per-family}
Tables~\ref{tab:per-family-ar} and~\ref{tab:per-family-diffusion} break the results down by model family, and gains vary across families:
\begin{itemize}[leftmargin=*]
    \item \method attains the highest Avg@4 and Best@4 on OLMo-3, Qwen3, E2D2, and BD3LM and the highest Best@4 on Marin, but falls behind GPT-6-sol-ICL in Avg@4 on Marin and E2D2-SW and behind Base-ICL in Best@4 on E2D2-SW.
    \item Although its AR advantage is largest on Qwen3, \method still achieves the highest AR Avg@4 and Best@4 when excluding Qwen3 (4.42 and 10.68, vs.\ at most 3.09 and 7.58).
    \item On NanoChat, headroom is limited: no method improves by more than about $2\sigma_e$ (Best@4 at most $1.23\times10^{-3}$ BPB), and GPT-6-sol-ICL is the only method with a positive Avg@4. 
\end{itemize}
These results show that \method can bring a smaller open-weight model to frontier-level performance in proposing effective LM improvements.


\subsection{Statistical Robustness}\label{app:robust}
We express experimental gains in units of task-specific seed noise $\sigma_e$, defined as the standard deviation of the reference configuration's validation loss over three runs.
Table~\ref{tab:sigma} reports Avg@4 and Best@4 in units of $\sigma_e$, the fraction of proposals with gains exceeding $2\sigma_e$, and Avg@4 relative to the reference loss.
On AR tasks, \method achieves an Avg@4 of $7.29\sigma_e$, compared with at most $4.91\sigma_e$ for the baselines.
Moreover, $50.8\%$ of its proposals exceed $2\sigma_e$, compared with $22.5\%$ for Base and $40.0\%$ for GPT-6-sol-ICL, while its Avg@4 corresponds to a $1.28\%$ reduction in reference loss.

\begin{table*}[t]
\centering
\small
\setlength{\tabcolsep}{3pt}

\caption{\textbf{Proposal-level results of Table~\ref{tab:main}, normalized by task-specific seed noise $\sigma_e$.}
Avg@4 and Best@4 are reported in units of $\sigma_e$; $>2\sigma_e$ denotes the fraction of proposals with gains exceeding $2\sigma_e$, and Rel.\ denotes Avg@4 as a percentage of reference loss.
Best results are in \textbf{bold}.}

\label{tab:sigma}
\resizebox{0.9\textwidth}{!}{
\begin{tabular}{l cccc cccc}
\toprule
& \multicolumn{4}{c}{\textbf{AR}} & \multicolumn{4}{c}{\textbf{Diffusion}} \\
\cmidrule(lr){2-5}\cmidrule(lr){6-9}
\textbf{Model} & Avg@4 $\uparrow$ & Best@4 $\uparrow$ & $>2\sigma_e$ (\%) $\uparrow$ & Rel. (\%) $\uparrow$ & Avg@4 $\uparrow$ & Best@4 $\uparrow$ & $>2\sigma_e$ (\%) $\uparrow$ & Rel. (\%) $\uparrow$ \\
\midrule
Base & $-8.83$ & $5.65$ & $22.5$ & $-0.75$ & $-31.61$ & $54.25$ & $34.7$ & $-1.25$ \\
Base-ICL & $0.73$ & $8.67$ & $30.8$ & $0.25$ & $28.10$ & $62.66$ & $73.6$ & $1.22$ \\
GPT-6-sol & $3.51$ & $4.94$ & $35.4$ & $0.47$ & $28.61$ & $43.02$ & $61.1$ & $0.60$ \\
GPT-6-sol-ICL & $4.91$ & $6.58$ & $40.0$ & $0.46$ & $33.99$ & $37.52$ & $\mathbf{100.0}$ & $1.49$ \\
SFT & $-11.84$ & $2.24$ & $11.3$ & $-1.01$ & $-11.71$ & $24.93$ & $31.9$ & $-0.90$ \\
RFT-Real & $-26.49$ & $4.85$ & $15.9$ & $-2.43$ & $-18.77$ & $36.96$ & $38.9$ & $-1.09$ \\
\method & $\mathbf{7.29}$ & $\mathbf{11.69}$ & $\mathbf{50.8}$ & $\mathbf{1.28}$ & $\mathbf{62.69}$ & $\mathbf{85.74}$ & $89.4$ & $\mathbf{2.15}$ \\
\bottomrule
\end{tabular}}
\end{table*}

\begin{table}[t]
\centering
\setlength{\tabcolsep}{3pt}
\caption{\textbf{Total cost}: data-generation cost (Table~\ref{tab:rq1-cost}) plus the GPU cost of proposer training, in B200 GPU-hours (H200 converted at a ratio of 1.198) and thousands of dollars at \$14.208 per GPU-hour.}
\label{tab:total-cost}
\coltablebox{
\begin{tabular}{ll ccc c}
\toprule
& & Data generation & \multicolumn{2}{c}{Proposer training} & Total \\
\cmidrule(lr){4-5}
Method & & $C_{\$}$ (\$k) & GPU-h & Cost (\$k) & (\$k) \\
\midrule
ICL ($K=100$--$300$) & & $19.8$--$60.0$ & $0$ & $0$ & $19.8$--$60.0$ \\
SFT & & $0$ & $1.6$ & $<0.1$ & $<0.1$ \\
RFT-Real & & $60.0$ & $0.8$ & $<0.1$ & $60.0$ \\
\midrule
\multirow{5}{*}{\method} & $M_0$ & $23.9$ & $152$ & $2.2$ & $26.1$ \\
 & $M_1$ & $28.3$ & $434$ & $6.2$ & $34.5$ \\
 & $M_2$ & $36.2$ & $898$ & $12.8$ & $48.9$ \\
 & $M_3$ & $40.1$ & $1{,}131$ & $16.1$ & $56.1$ \\
 & $M_4$ & $46.6$ & $1{,}376$ & $19.6$ & $66.2$ \\
\bottomrule
\end{tabular}}
\end{table}

\section{Cost Analysis}\label{app:cost}
\subsection{Data-Generation Cost}\label{app:data-cost}
\begin{figure*}[t]
\centering
\begin{subfigure}[t]{0.24\textwidth}
\centering
\includegraphics[width=\linewidth]{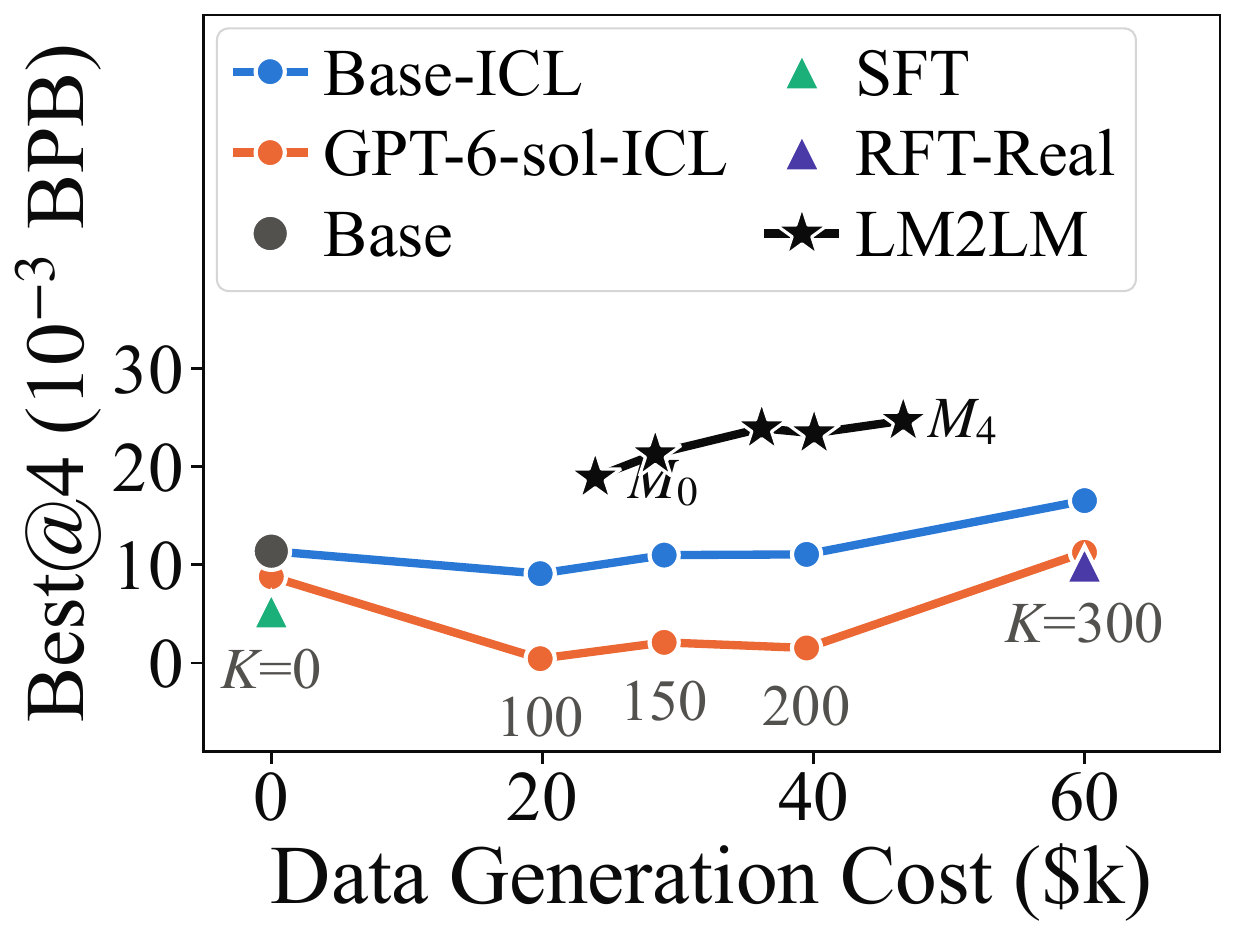}
\caption{AR: Best@4 vs.\ cost.}
\label{fig:cost-best-ar-cost}
\end{subfigure}\hfill
\begin{subfigure}[t]{0.24\textwidth}
\centering
\includegraphics[width=\linewidth]{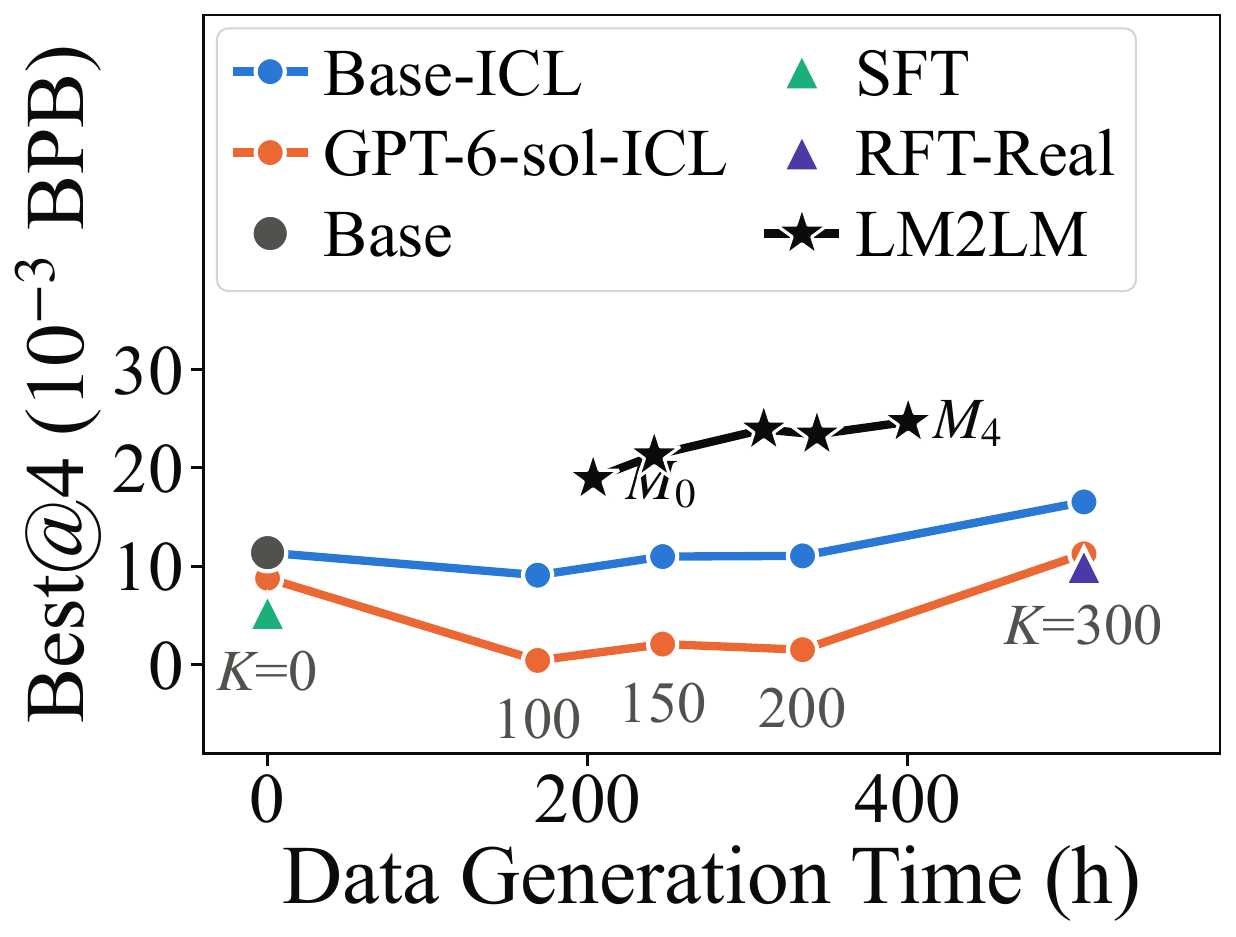}
\caption{AR: Best@4 vs.\ time.}
\label{fig:cost-best-ar-time}
\end{subfigure}\hfill
\begin{subfigure}[t]{0.24\textwidth}
\centering
\includegraphics[width=\linewidth]{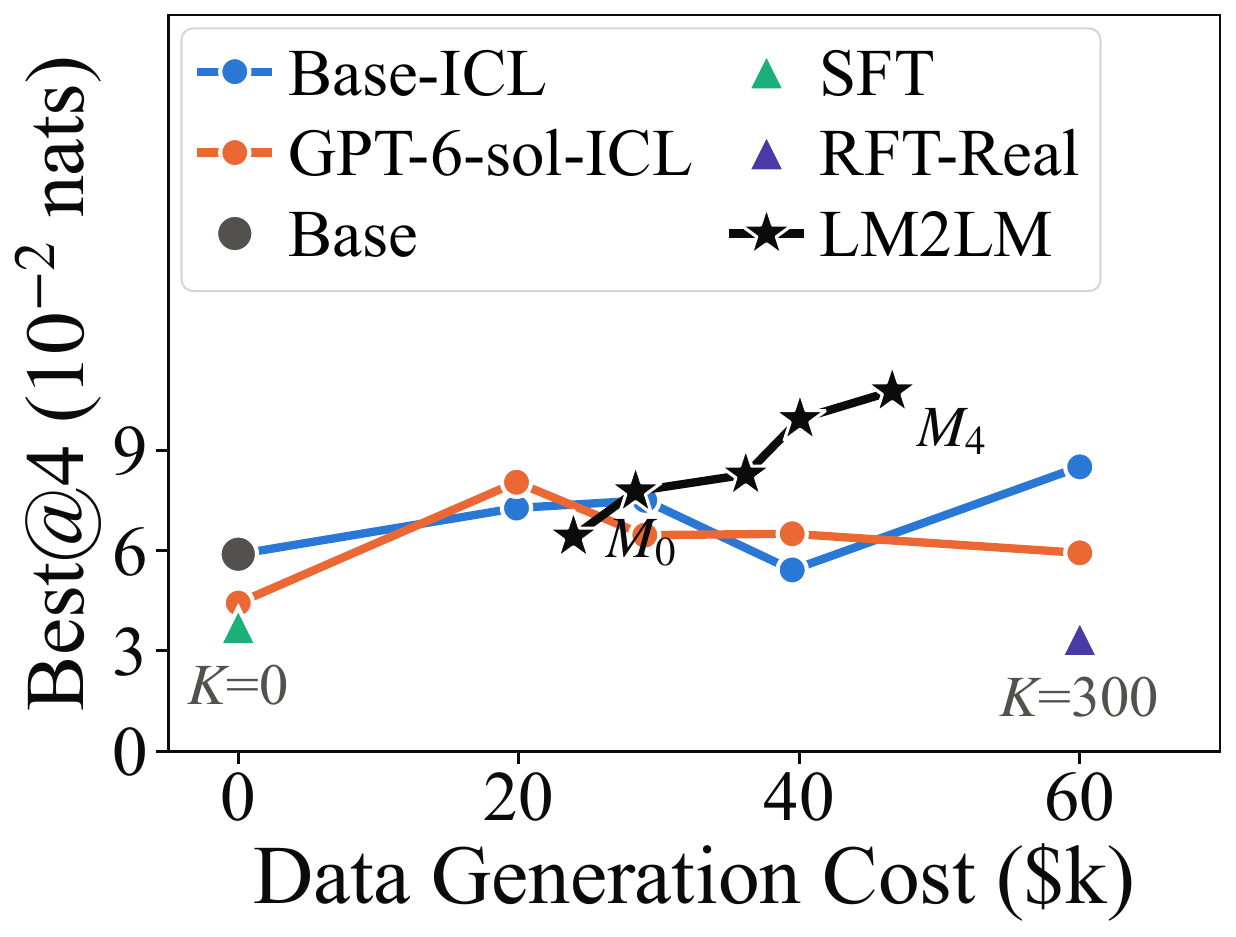}
\caption{Diff.: Best@4 vs.\ cost.}
\label{fig:cost-best-diff-cost}
\end{subfigure}\hfill
\begin{subfigure}[t]{0.24\textwidth}
\centering
\includegraphics[width=\linewidth]{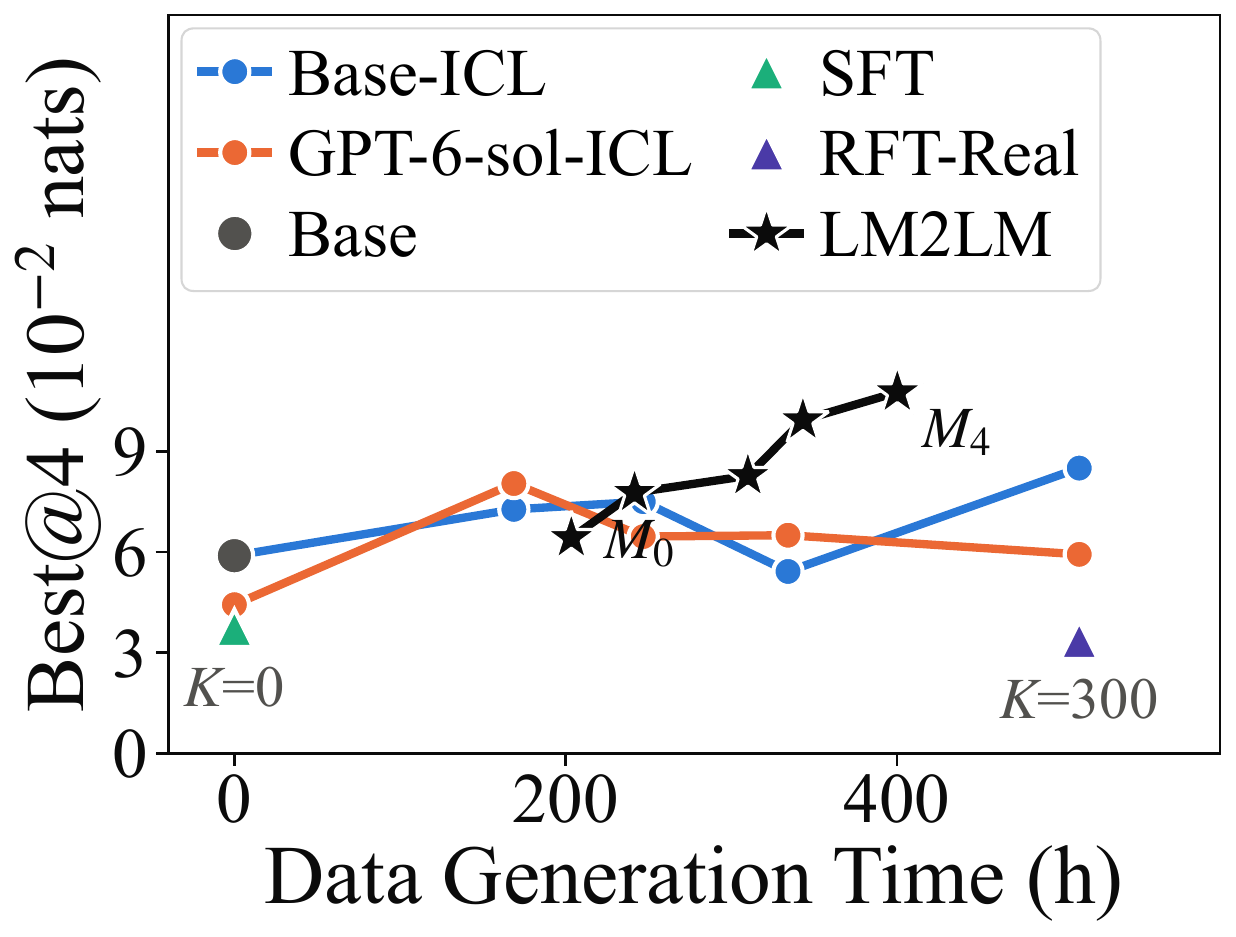}
\caption{Diff.: Best@4 vs.\ time.}
\label{fig:cost-best-diff-time}
\end{subfigure}
\caption{\textbf{Real experimental gains (Best@4) versus data-generation cost and time on held-out research states.}
(a, b) show AR tasks in $10^{-3}$ BPB, while (c, d) show diffusion tasks in $10^{-2}$ nats per token.}
\label{fig:cost-best}
\end{figure*}

Fig.~\ref{fig:cost-best} reports Best@4 against data-generation cost and time, complementing the Avg@4 results in Fig.~\ref{fig:cost}; Table~\ref{tab:rq1-cost} lists the underlying numbers.
The trends match those of Avg@4: on AR tasks, \method exceeds every baseline already before its first calibration (18.91 at \$23.9k, vs.\ at most 16.54 at \$60.0k) and saturates after the second calibration, whereas on diffusion tasks its Best@4 rises with each calibration and exceeds every baseline at $M_3$ (9.93 vs.\ at most 8.50).

\begin{table*}[t]
\centering\small
\setlength{\tabcolsep}{4pt}

\caption{\textbf{Data-generation cost, time, and real experimental gains on held-out AR states} (Fig.~\ref{fig:cost}).
$T_{\mathrm{run}}$ and $T_{\mathrm{RWM}}$ denote cumulative real-execution and RWM prediction time.
Gains are reported in $10^{-3}$ BPB.}
\label{tab:rq1-cost}
\resizebox{0.7\textwidth}{!}{
\begin{tabular}{ll cccccc cc}
\toprule
& & \multicolumn{6}{c}{Data-generation cost} & \multicolumn{2}{c}{Gain $\uparrow$} \\
\cmidrule(lr){3-8}\cmidrule(lr){9-10}
Method & $K$ & Real runs & GPU-h & Tokens (M) & $C_{\$}$ & $T_{\mathrm{run}}$ (h) & $T_{\mathrm{RWM}}$ (h) & Avg@4 & Best@4 \\
\midrule
\multirow{5}{*}{Base-ICL} & $0$ & $0$ & $0$ & $0.0$ & $0$ & $0.0$ & $0.0$ & $-7.94$ & $11.39$ \\
 & $100$ & $100$ & $1{,}396$ & $0.0$ & $19{,}839$ & $168.7$ & $0.0$ & $-7.64$ & $9.10$ \\
 & $150$ & $150$ & $2{,}040$ & $0.0$ & $28{,}986$ & $246.8$ & $0.0$ & $-2.12$ & $10.99$ \\
 & $200$ & $200$ & $2{,}780$ & $0.0$ & $39{,}495$ & $334.2$ & $0.0$ & $-1.98$ & $11.06$ \\
 & $300$ & $300$ & $4{,}223$ & $0.0$ & $60{,}001$ & $510.0$ & $0.0$ & $4.04$ & $16.54$ \\
\midrule
\multirow{5}{*}{GPT-6-sol-ICL} & $0$ & $0$ & $0$ & $0.0$ & $0$ & $0.0$ & $0.0$ & $6.51$ & $8.79$ \\
 & $100$ & $100$ & $1{,}396$ & $0.0$ & $19{,}839$ & $168.7$ & $0.0$ & $-1.99$ & $0.44$ \\
 & $150$ & $150$ & $2{,}040$ & $0.0$ & $28{,}986$ & $246.8$ & $0.0$ & $-0.56$ & $2.10$ \\
 & $200$ & $200$ & $2{,}780$ & $0.0$ & $39{,}495$ & $334.2$ & $0.0$ & $-0.70$ & $1.53$ \\
 & $300$ & $300$ & $4{,}223$ & $0.0$ & $60{,}001$ & $510.0$ & $0.0$ & $6.15$ & $11.27$ \\
\midrule
SFT & $300$ & $0$ & $0$ & $0.0$ & $0$ & $0.0$ & $0.0$ & $-12.19$ & $5.51$ \\
RFT-Real & $300$ & $300$ & $4{,}223$ & $0.0$ & $60{,}001$ & $510.0$ & $0.0$ & $-22.95$ & $9.69$ \\
\midrule
\multirow{5}{*}{\method} & $M_0$ & $100$ & $1{,}665$ & $140.6$ & $23{,}901$ & $202.0$ & $1.4$ & $1.81$ & $18.91$ \\
 & $M_1$ & $131$ & $1{,}960$ & $294.1$ & $28{,}330$ & $239.0$ & $2.5$ & $10.85$ & $21.28$ \\
 & $M_2$ & $169$ & $2{,}500$ & $459.4$ & $36{,}183$ & $306.6$ & $3.4$ & $18.01$ & $23.91$ \\
 & $M_3$ & $200$ & $2{,}758$ & $664.7$ & $40{,}050$ & $338.9$ & $4.3$ & $17.06$ & $23.40$ \\
 & $M_4$ & $236$ & $3{,}206$ & $889.3$ & $46{,}626$ & $395.0$ & $5.3$ & $\mathbf{19.40}$ & $\mathbf{24.69}$ \\
\bottomrule
\end{tabular}}
\end{table*}

\subsection{Total Cost}\label{app:total-cost}
We further account for proposer training costs beyond data generation to assess the overall cost-effectiveness of \method.

Table~\ref{tab:total-cost} reports the total costs, including RL training for \method and fine-tuning for SFT and RFT-Real, while ICL requires no training.
At step 96 ($M_3$), \method incurs \$40.1k in data-generation costs and \$16.1k in RL training costs, totaling \$56.1k.
Despite its additional RL training cost, \method remains cheaper than ICL with $K=300$ (\$60.0k), while achieving substantially higher Avg@4 on AR (17.06 vs.\ 6.15) and diffusion tasks (7.10 vs.\ 4.94).

\begin{figure}[t]
\centering
\begin{subfigure}[t]{0.49\colfigwidth}\centering\includegraphics[width=\linewidth]{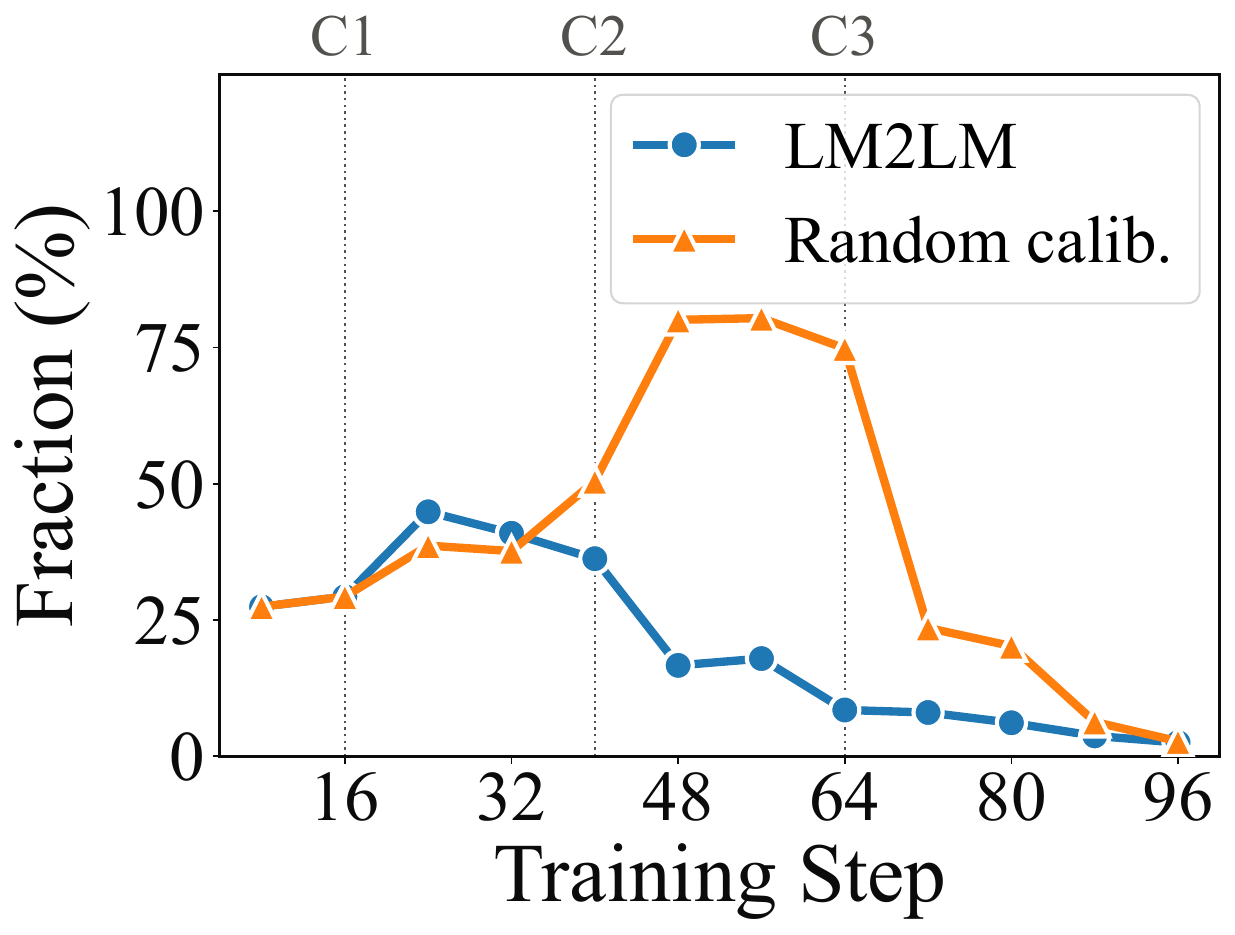}\caption{Training share.}\label{fig:case-calib-share}\end{subfigure}\hspace{0.02\colfigwidth}
\begin{subfigure}[t]{0.49\colfigwidth}\centering\includegraphics[width=\linewidth]{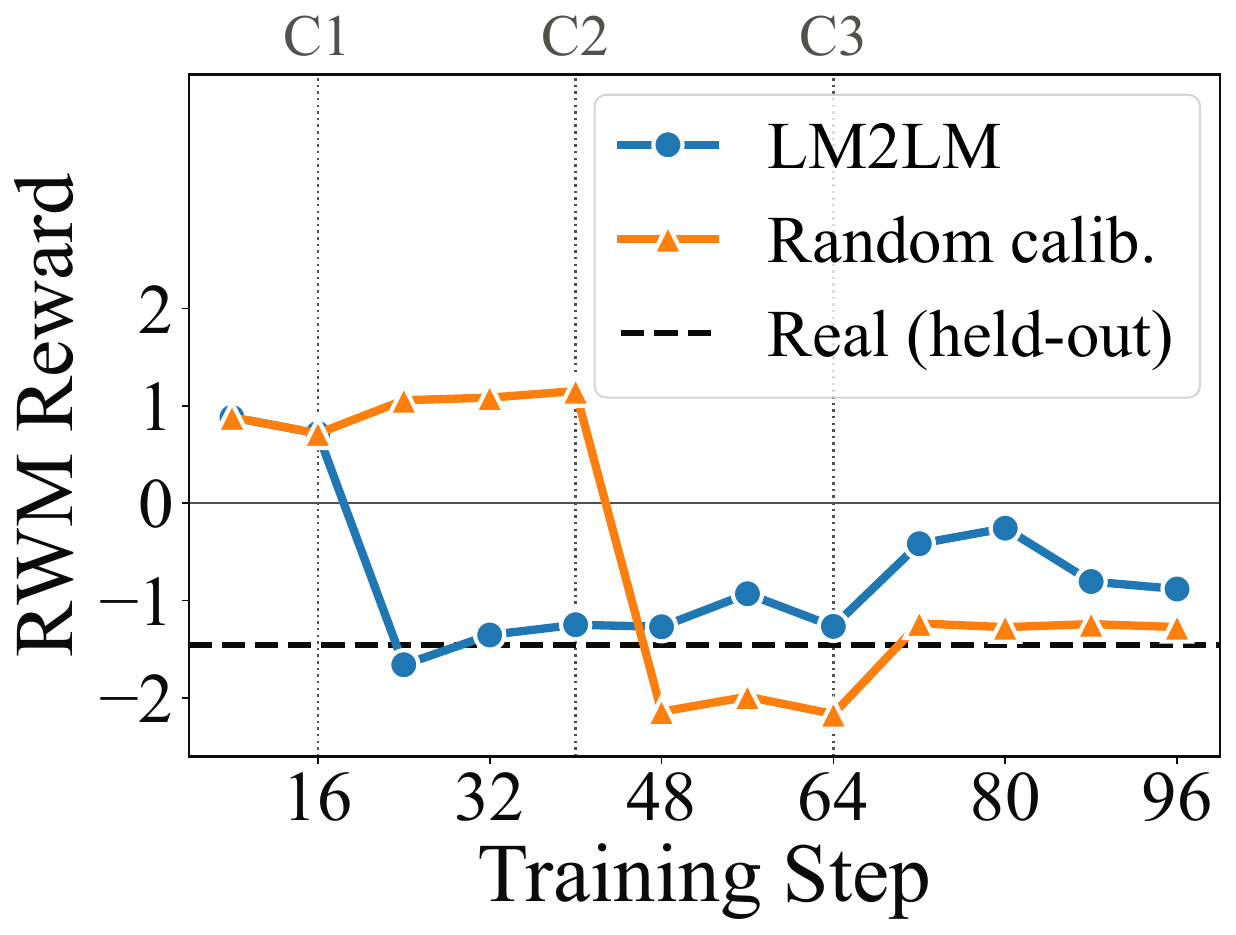}\caption{RWM reward.}\label{fig:case-calib-reward}\end{subfigure}
\caption{\textbf{M-UPD-72 under top-rated and random calibration.} (a) Fraction of NanoChat training proposals that apply it. (b) Mean RWM reward of these proposals in the training reward scale; the dashed line is their real reward on held-out states. C1--C3 mark the three calibrations.}
\label{fig:case-calib}
\end{figure}

\begin{figure}[t]
\centering
\begin{subfigure}[t]{0.49\colfigwidth}\centering\includegraphics[width=\linewidth]{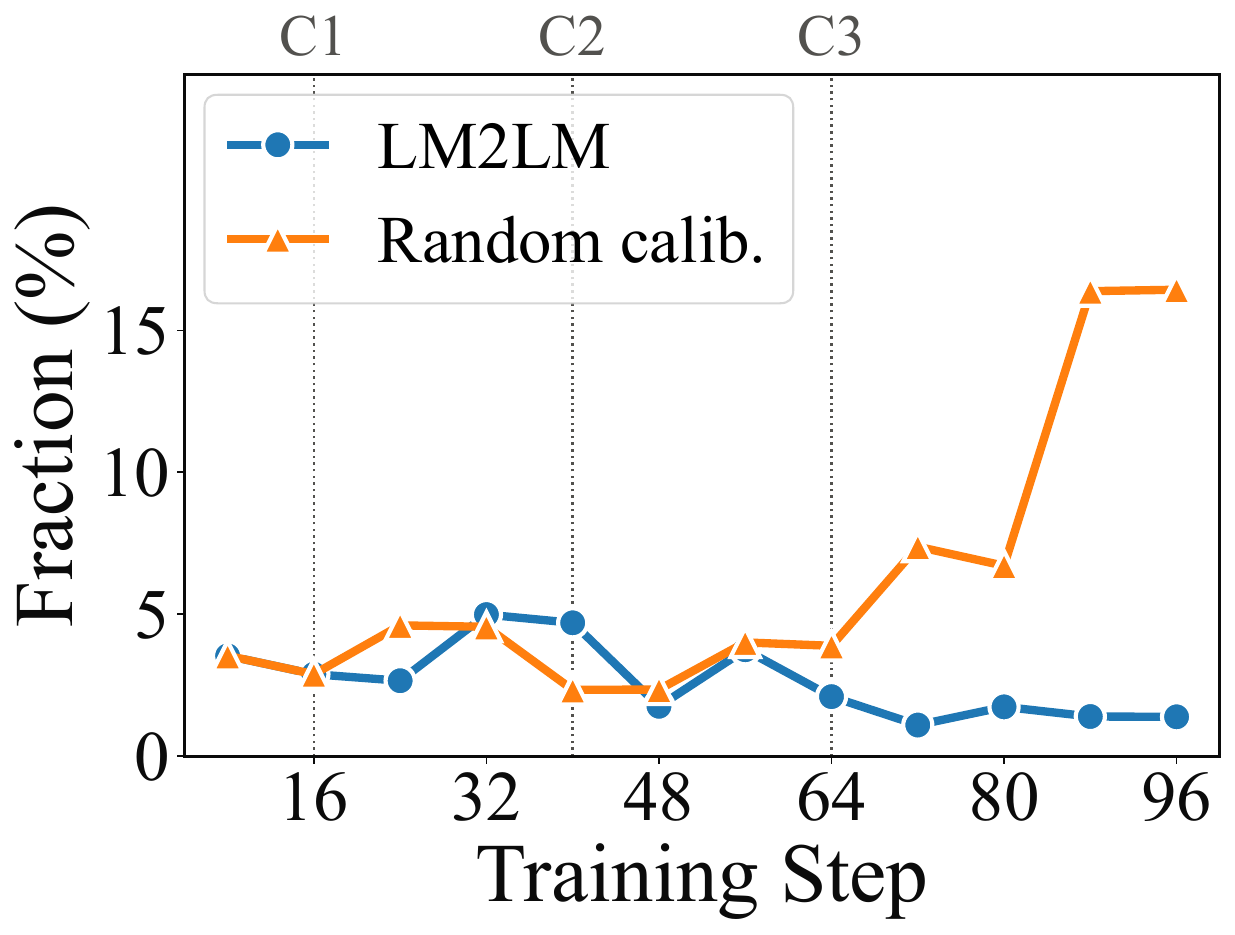}\caption{Training share.}\label{fig:case-tokenmixer-share}\end{subfigure}\hspace{0.02\colfigwidth}
\begin{subfigure}[t]{0.49\colfigwidth}\centering\includegraphics[width=\linewidth]{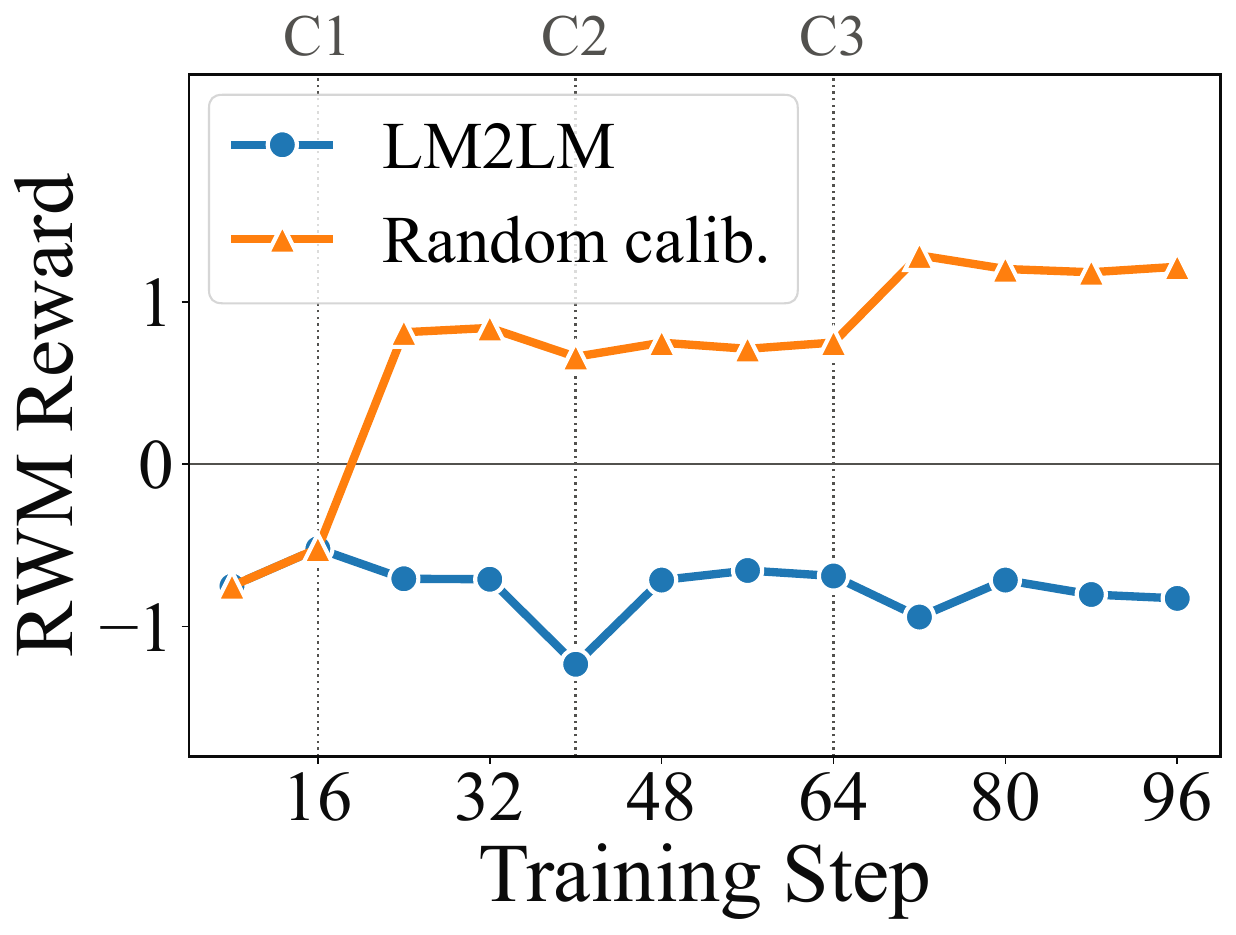}\caption{RWM reward.}\label{fig:case-tokenmixer-reward}\end{subfigure}
\caption{\textbf{OLMo token-mixing change under top-rated and random calibration.} (a) Fraction of OLMo training proposals that apply it. (b) Mean RWM reward of these proposals in the training reward scale. C1--C3 mark the three calibrations.}
\label{fig:case-tokenmixer}
\end{figure}

\section{Case Studies: Impact of Calibration Strategies}\label{app:ablation}
Top-rated calibration executes the proposals that the RWM rates highest and RL is about to reinforce, whereas random calibration samples proposals regardless of their predicted gains.
The following two cases illustrate how this difference affects RL on held-out AR families.

\noindent\textbf{Delayed Correction of Overestimated Changes.}
M-UPD-72 is a NanoChat change to the number of tokens per optimizer update that the RWM overestimates: under $M_0$, it rewards the change with $+0.7$ to $+0.9$ in the training reward scale, whereas its real reward on held-out states is about $-1.46$ (Fig.~\ref{fig:case-calib}).
Top-rated calibration executes seven proposals applying it in the first update (after step 16), five of which degrade the reference configuration; the RWM then rewards the change with $-1.66$, and its share of NanoChat training proposals falls from a peak of 45\% at step 24 to 8\% by step 64.
Random calibration executes none of them in the first update, so the RWM keeps rewarding the change until the second update (after step 40); lagging behind the reward, its share still peaks at 80\% at step 48 and falls to 24\% only by step 72.
Both shares eventually drop below 3\% by step 96, but random calibration spends substantially more training on a harmful change before correcting it.

\noindent\textbf{Misleading Records from Random Selection.}
Random calibration can also add records that mislead the RWM.
Its first batch contains an OLMo change that replaces local attention with an alternative token-mixing operator, which happens to improve its training task by 1.43\%.
With this record, the RWM starts rewarding the change (mean training reward rising from about $-0.6$ under $M_0$ to $+0.8$ after the first update and $+1.2$ by step 96), and RL raises its share of OLMo training proposals from 4\% to 16\% (Fig.~\ref{fig:case-tokenmixer}).
On held-out OLMo states, the change accounts for 53\% of the proposals at step 96, with a mean real gain of $-16.4\times10^{-3}$ BPB, lowering the OLMo Avg@4 from 7.39 under top-rated calibration to $-6.37$.
Under top-rated calibration, the RWM keeps penalizing the change ($-1.2$ to $-0.5$), and its share stays below 5\%.


\section{Additional In-depth Analysis}\label{app:analysis}

\subsection{Diagnosing RWM Feedback Reliability}\label{app:feedback-metrics}\label{app:feedback-diagnostics}
\noindent\textbf{Feedback Metric.} 
At each checkpoint, we execute the four proposals of each of the 60 held-out AR states, whose outcomes never enter the records.
The static variant is scored by $W(M_0)$ throughout, and \method by the RWM in force: $W(M_0)$ at step 16, $W(M_1)$ at step 40, $W(M_2)$ at steps 48 and 64, $W(M_3)$ at step 96, and $W(M_4)$ at step 128.
Rewards follow the normalization and clipping used for RL training, $\mathrm{clip}(g/\tau_{\mathrm{eff}},\pm 6)$.
For the proposals $i=1,\dots,G$ of a state, let $\hat r_i$ and $r_i$ denote their predicted and real rewards, and let
\[
\hat A_i=\hat r_i-\frac{1}{G}\sum_{j=1}^{G}\hat r_j,\qquad A_i=r_i-\frac{1}{G}\sum_{j=1}^{G}r_j
\]
be their predicted and real advantages, with $G=4$ (states with fewer than two scored proposals are skipped).
We use the following metrics, with 95\% bootstrap intervals over states (2{,}000 resamples):
\begin{itemize}[leftmargin=*]
    \item \emph{Advantage cosine}: the cosine similarity between the predicted and real advantages, concatenated over all states.
    \item \emph{Pairwise accuracy}: over all pairs of proposals of a state whose real rewards differ, the fraction whose predicted order agrees with their real order, where a pair with equal predicted rewards counts as 0.5; pairs are pooled over states, so random rewards attain 0.5.
    \item \emph{Sign agreement}: the fraction of proposals with nonzero $\hat A_i$ and $A_i$ whose signs agree.
    \item \emph{Wrong-reinforcement rate}: $\Pr(A_i<0\mid \hat A_i>0)$, the fraction of positively reinforced proposals that perform below their group's real mean.
    \item \emph{Top-1 hit rate}: over states whose real rewards differ, the fraction in which the proposal with the highest predicted reward also has the highest real reward.
    \item \emph{Real advantage of reinforced proposals}: the mean $A_i$ over proposals with $\hat A_i>0$.
\end{itemize}
We also report the mean predicted and real rewards and their difference (Fig.~\ref{fig:adv}(c)).

\begin{figure*}[t]
\centering
\begin{subfigure}[t]{0.24\textwidth}\centering\includegraphics[width=\linewidth]{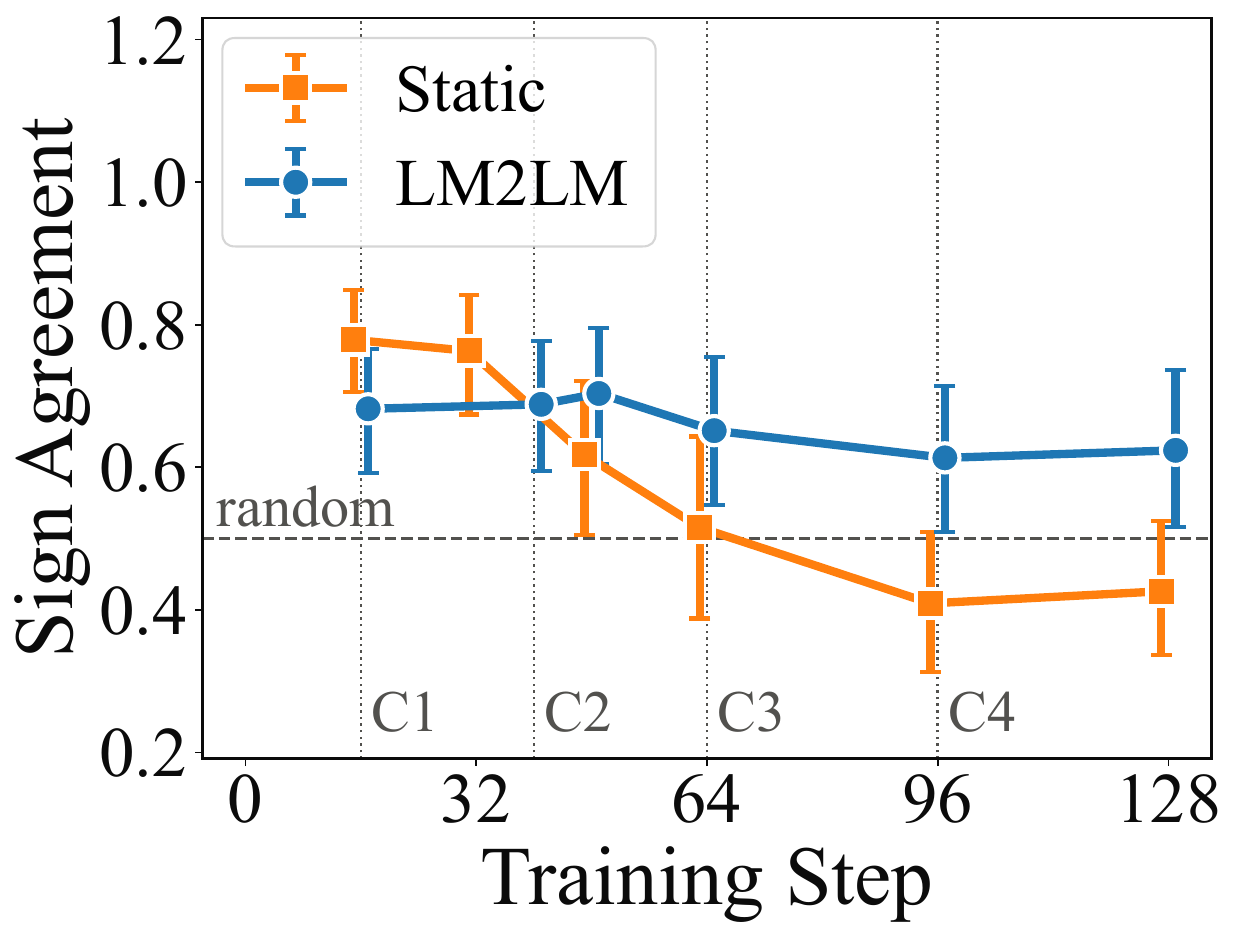}\caption{Sign agreement.}\end{subfigure}\hfill
\begin{subfigure}[t]{0.24\textwidth}\centering\includegraphics[width=\linewidth]{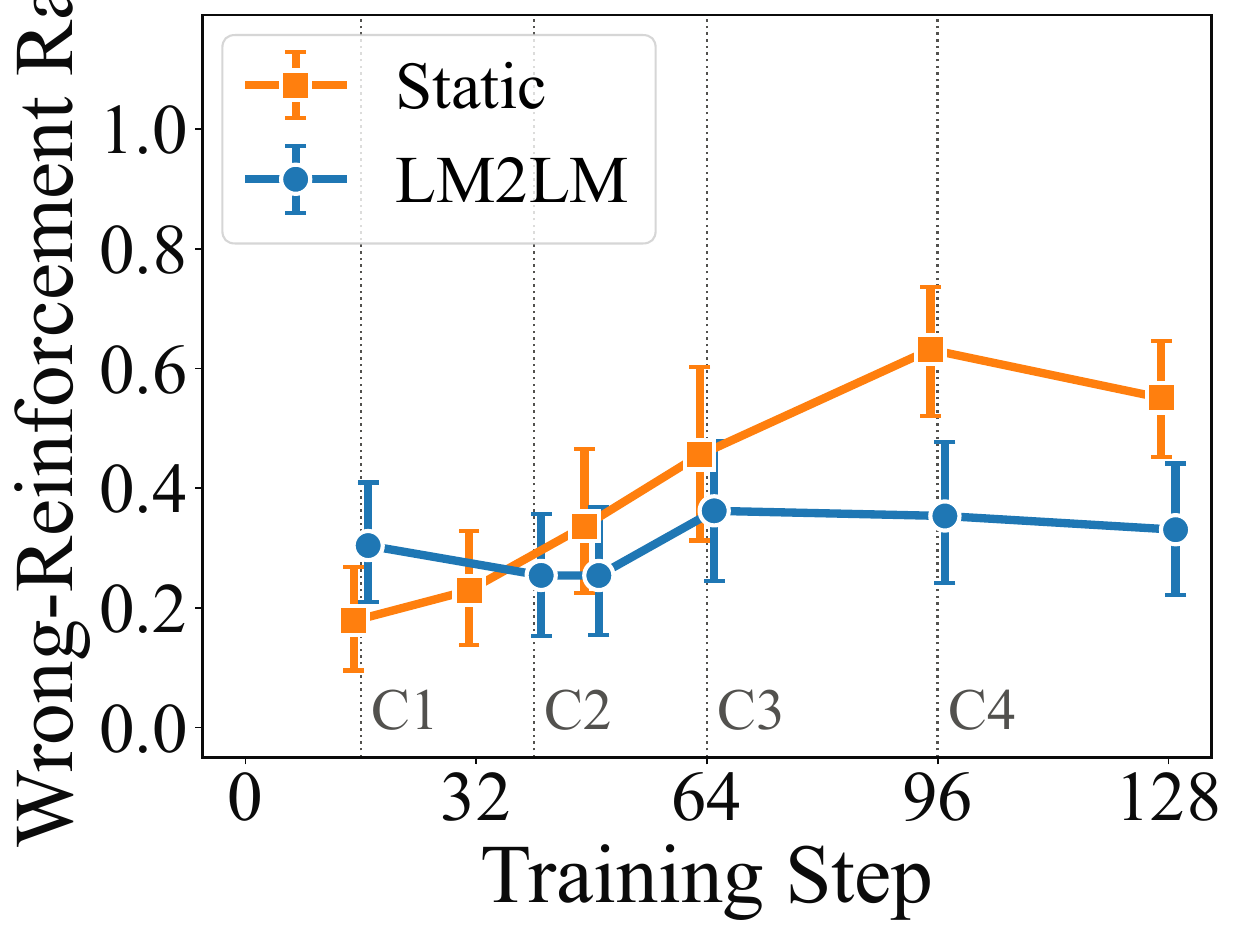}\caption{Wrong reinforcement.}\end{subfigure}\hfill
\begin{subfigure}[t]{0.24\textwidth}\centering\includegraphics[width=\linewidth]{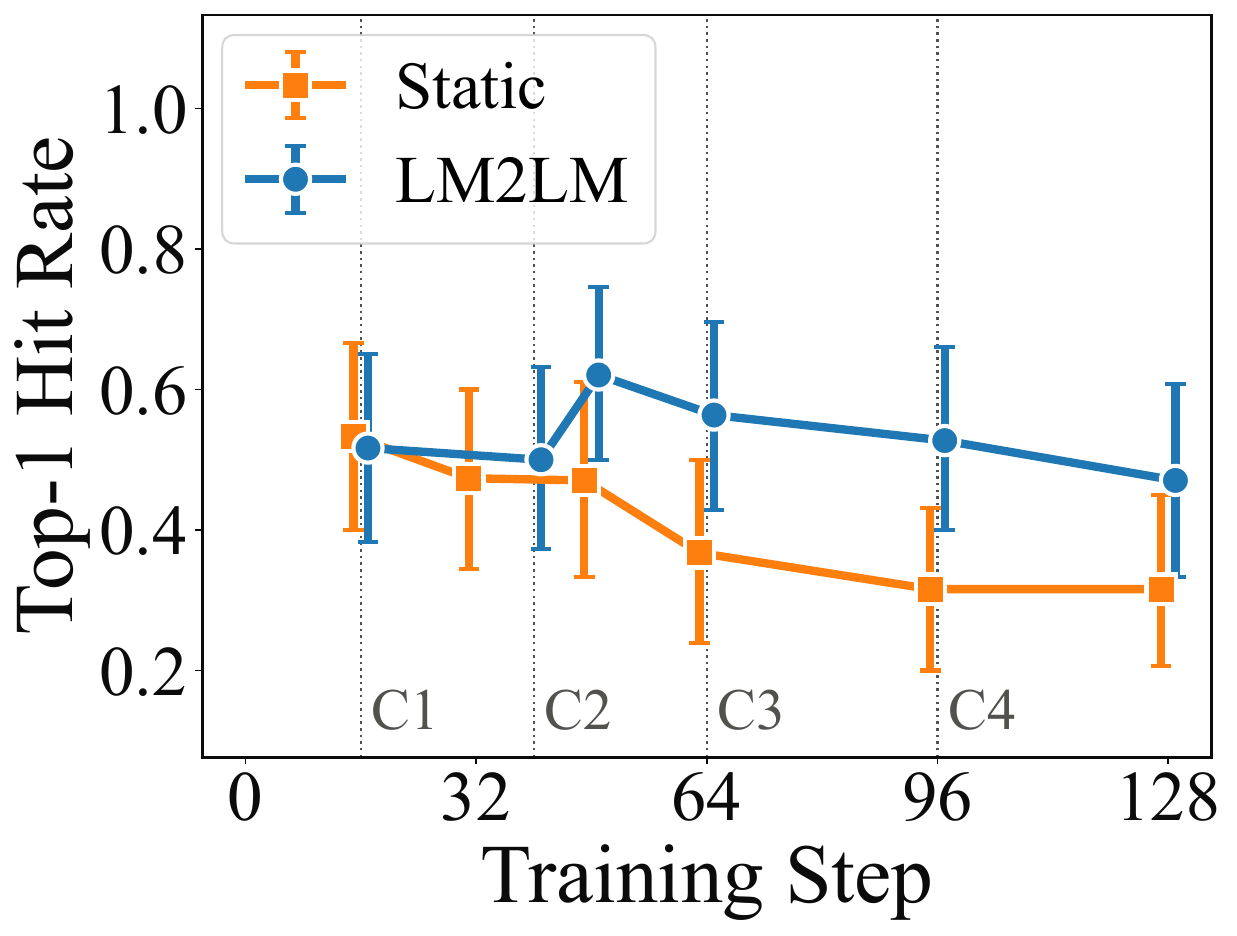}\caption{Top-1 hit rate.}\end{subfigure}\hfill
\begin{subfigure}[t]{0.24\textwidth}\centering\includegraphics[width=\linewidth]{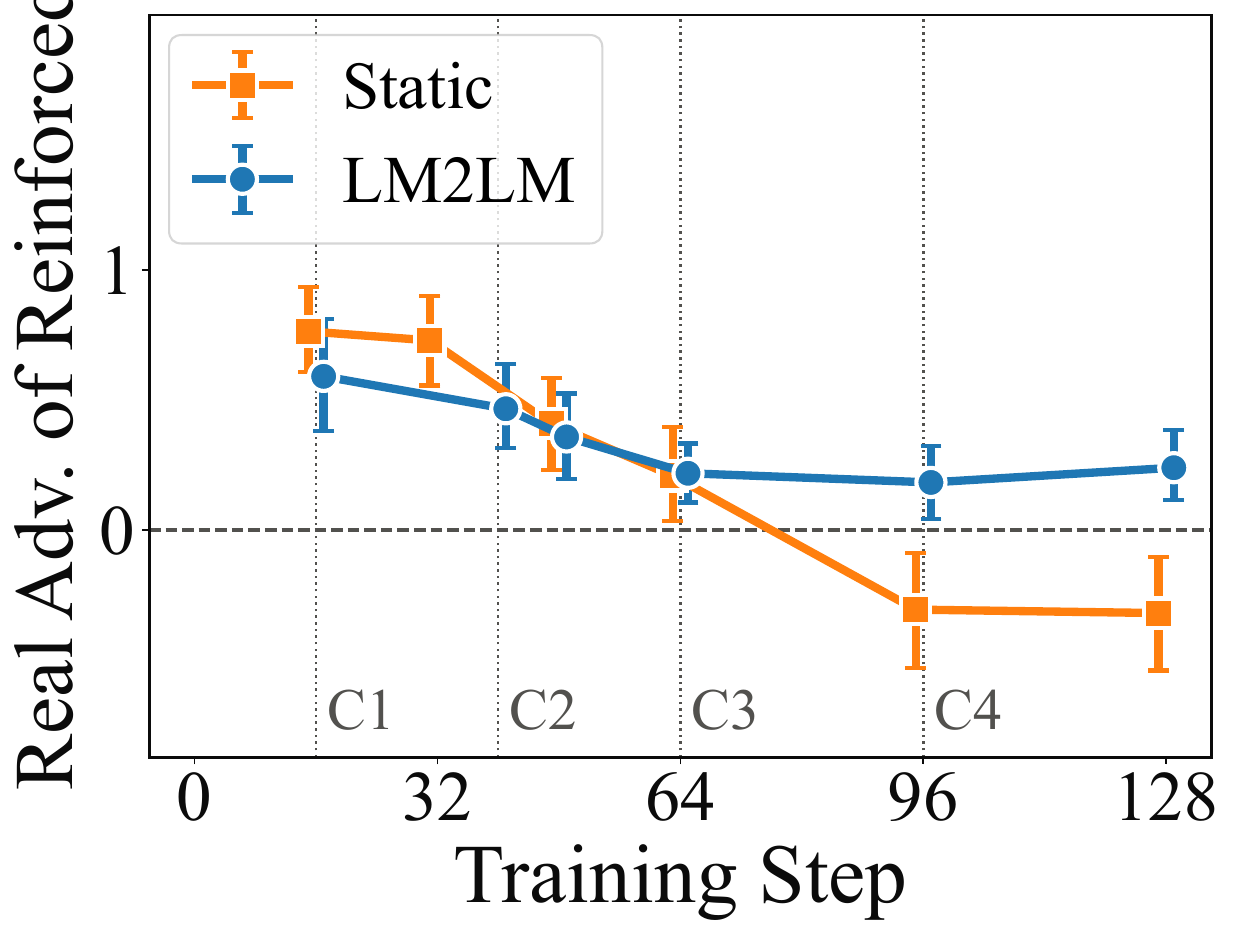}\caption{Real advantage of reinforced.}\end{subfigure}
\caption{\textbf{Additional feedback diagnostics on held-out AR states through step 128.} Each variant is evaluated using its own training RWM. (d) reports the mean real advantage of proposals with positive predicted advantage. Error bars indicate 95\% bootstrap confidence intervals over states.}
\label{fig:app-diag}
\end{figure*}

\begin{figure*}[t]
\centering
\includegraphics[width=0.6\textwidth]{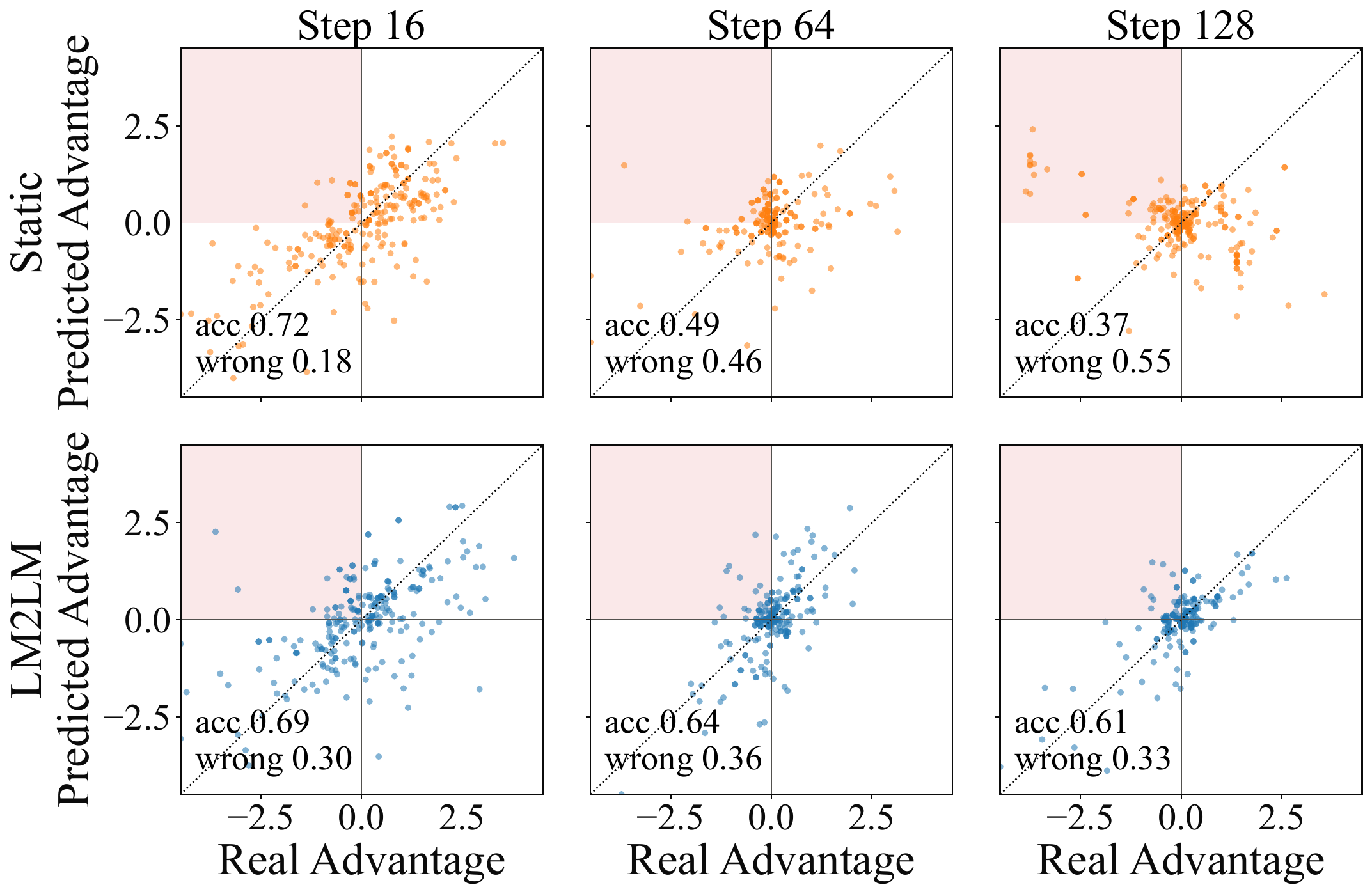}
\caption{\textbf{Predicted versus real advantages of the proposer's own proposals on held-out AR states, at steps 16, 64, and 128.} Top: static variant scored by $W(M_0)$; bottom: \method scored by the RWM in force. Points on the dotted diagonal are predicted exactly; the shaded region contains wrongly reinforced proposals (positive predicted, negative real advantage). Each panel reports the pairwise accuracy (acc) and wrong-reinforcement rate (wrong). Values are clipped to $\pm 4.5$ for display.}
\label{fig:adv-scatter}
\end{figure*}

\noindent\textbf{Results.}
Fig.~\ref{fig:app-diag} complements the advantage alignment and ranking results of Sec.~\ref{ssec:RQ2} by examining whether individual proposals are reinforced in the correct direction.
Early in training, the static RWM provides feedback comparable to that of \method, but its reliability deteriorates as the proposer evolves.
From step 16 to step 64, its sign agreement drops from 0.78 to 0.52 and its wrong-reinforcement rate rises from 18\% to 46\%, reaching 0.43 and 55\% at step 128.
In contrast, the sign agreement of \method is 0.68, 0.65, and 0.62 at steps 16, 64, and 128, and its wrong-reinforcement rate stays between 30\% and 36\% at these steps.
As a result, the proposals that the static RWM reinforces have a negative mean real advantage from step 96 onward ($-0.31$), i.e., RL increasingly favors proposals that underperform their group, whereas those reinforced by \method remain positive (0.18 to 0.59).
Fig.~\ref{fig:adv-scatter} shows the same trend at the level of individual proposals.


\subsection{Effect of Updating Records on Identical Proposals}\label{app:same-proposals}
\begin{figure}[t]
\centering
\includegraphics[width=0.7\colfigwidth]{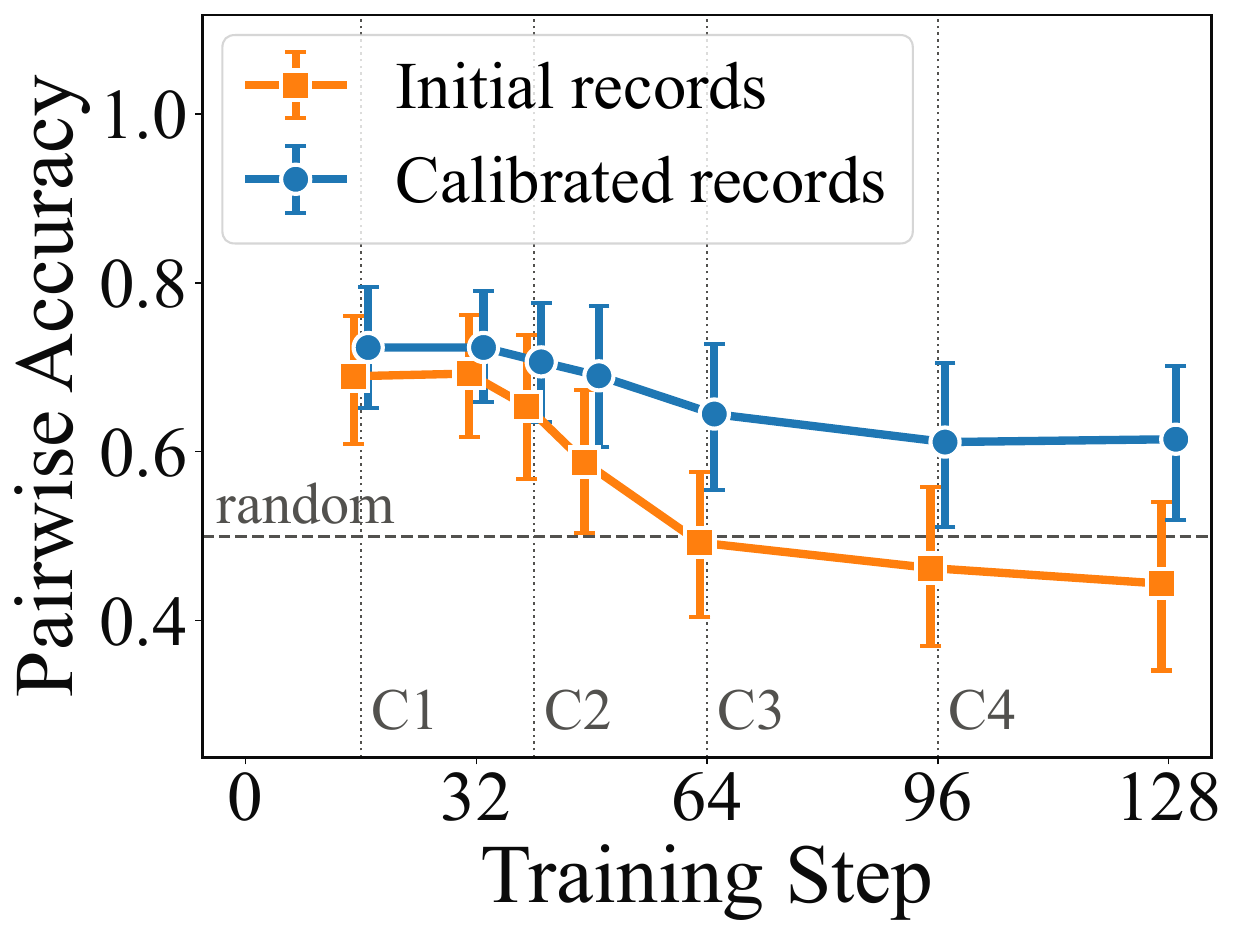}
\caption{\textbf{Pairwise accuracy of the same proposals} of \method scored with the initial records $M_0$ and with the latest calibrated records ($M_2$ up to step 64, $M_3$ at step 96, and $M_4$ at step 128).}
\label{fig:app-same}
\end{figure}

The comparison in Sec.~\ref{ssec:RQ2} involves two proposers that diverge during training.
To isolate the effect of the records, we score the same proposals of \method with the initial records $M_0$ and with the calibrated records.
With $M_0$, the pairwise accuracy drops to 0.49 at step 64, 0.46 at step 96, and 0.44 at step 128, whereas with the calibrated records it is 0.64 with $M_2$ at step 64, 0.61 with $M_3$ at step 96, and 0.61 with $M_4$ at step 128 (Fig.~\ref{fig:app-same}).
Since the proposals are identical, these differences come from the records alone.

\subsection{Case Studies}\label{app:feedback-cases}
\noindent\textbf{Family-dependent optimizer changes.}
The RWM rewards an orthogonalized-momentum optimizer with $+1.0$ to $+2.0$ on OLMo-3 and Qwen3 training tasks but with $-2$ to $-3$ on NanoChat.
The proposer follows this feedback: the share of the change in its OLMo-3 proposals rises from 7\% to 78\%, while on NanoChat it drops to 0\%.
Real execution confirms both decisions: on held-out states, the change improves 93\% of the OLMo-3 and 100\% of the Qwen3 proposals that apply it, by up to $21.6\times10^{-3}$ BPB on a single OLMo-3 state, whereas on NanoChat, where Base proposes it, it loses $44\times10^{-3}$ BPB on average.

\begin{figure*}[t]
\centering
\begin{subfigure}[t]{0.32\textwidth}\centering\includegraphics[width=\linewidth]{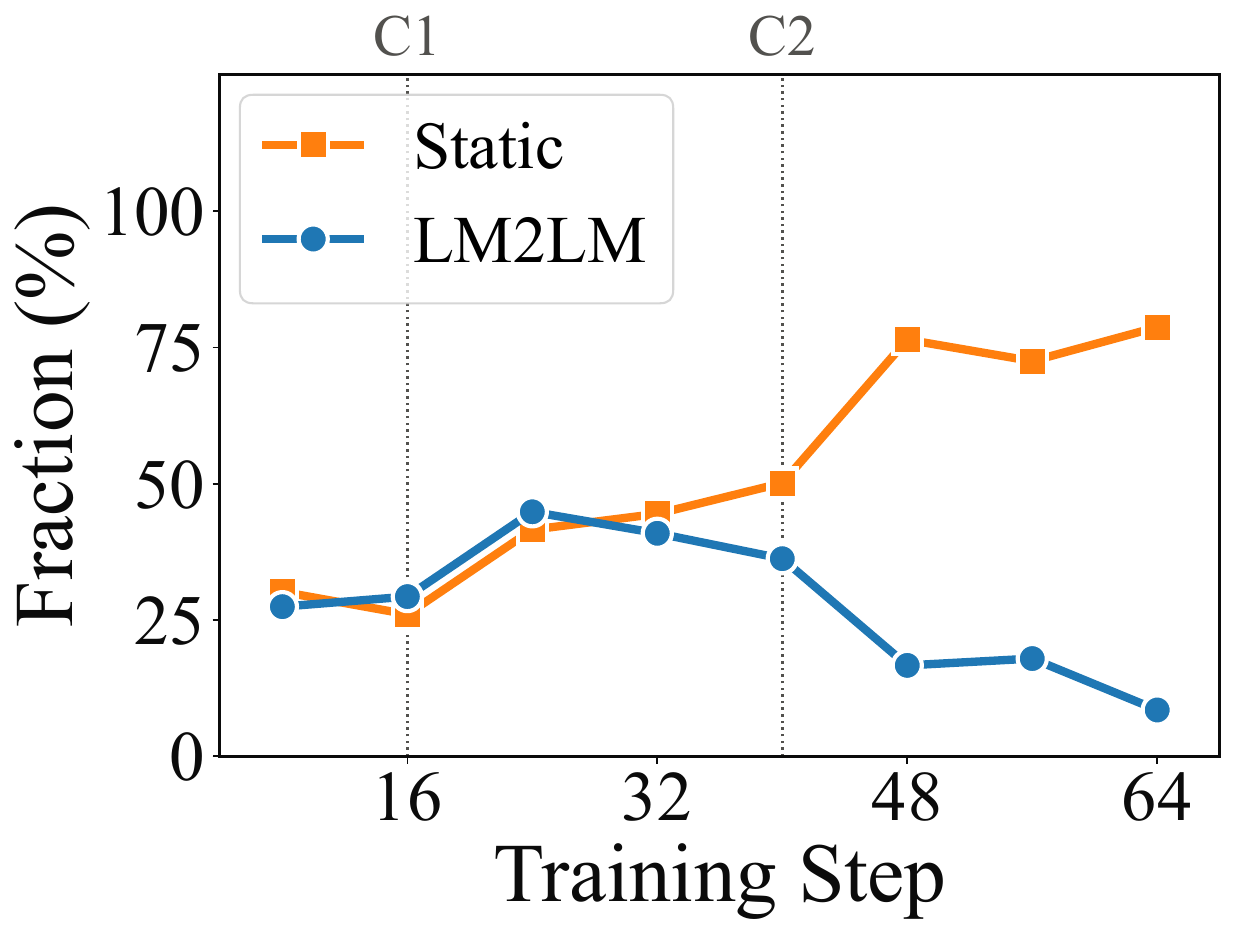}\caption{Training share.}\label{fig:case-mupd72-share}\end{subfigure}\hfill
\begin{subfigure}[t]{0.32\textwidth}\centering\includegraphics[width=\linewidth]{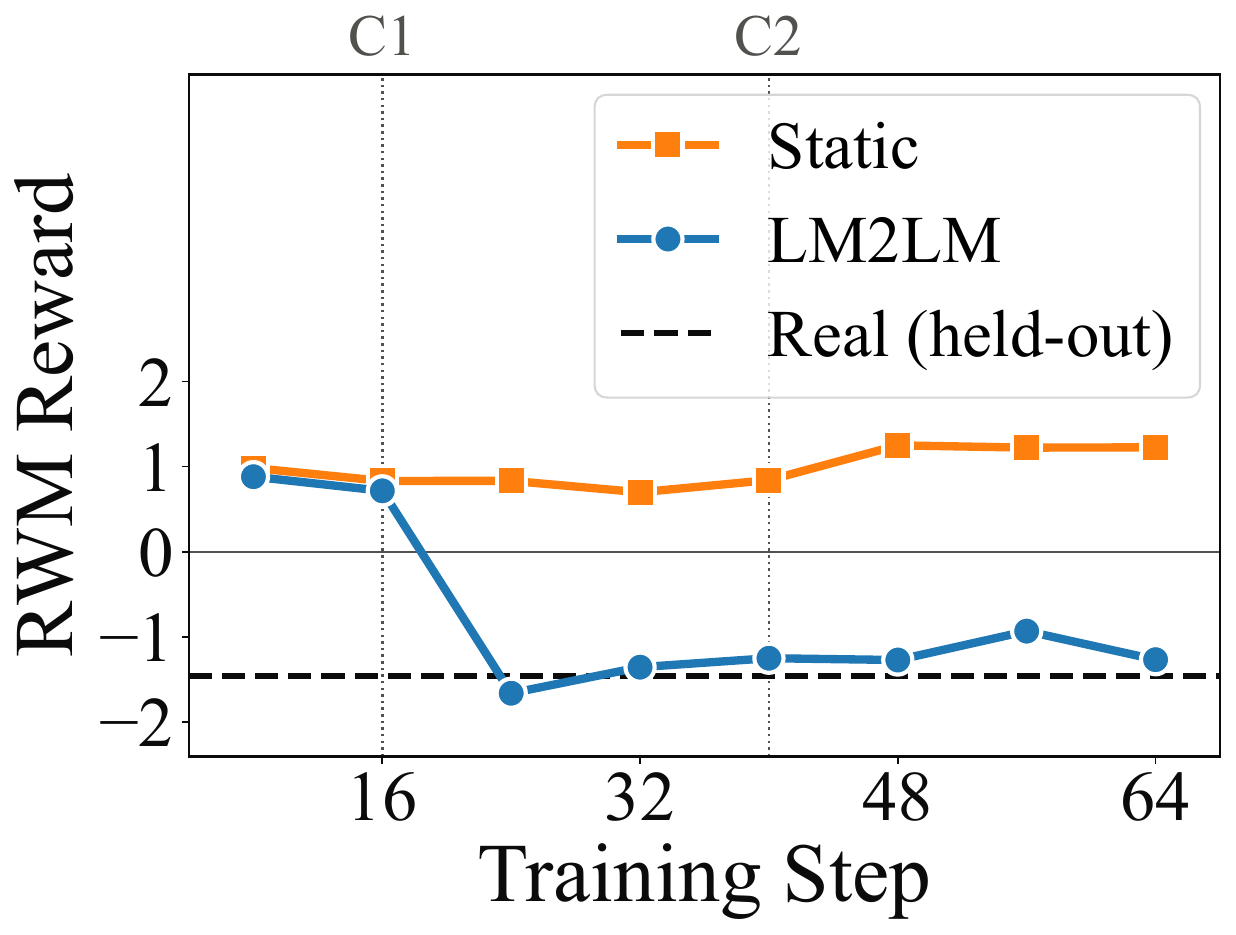}\caption{RWM reward.}\label{fig:case-mupd72-reward}\end{subfigure}\hfill
\begin{subfigure}[t]{0.32\textwidth}\centering\includegraphics[width=\linewidth]{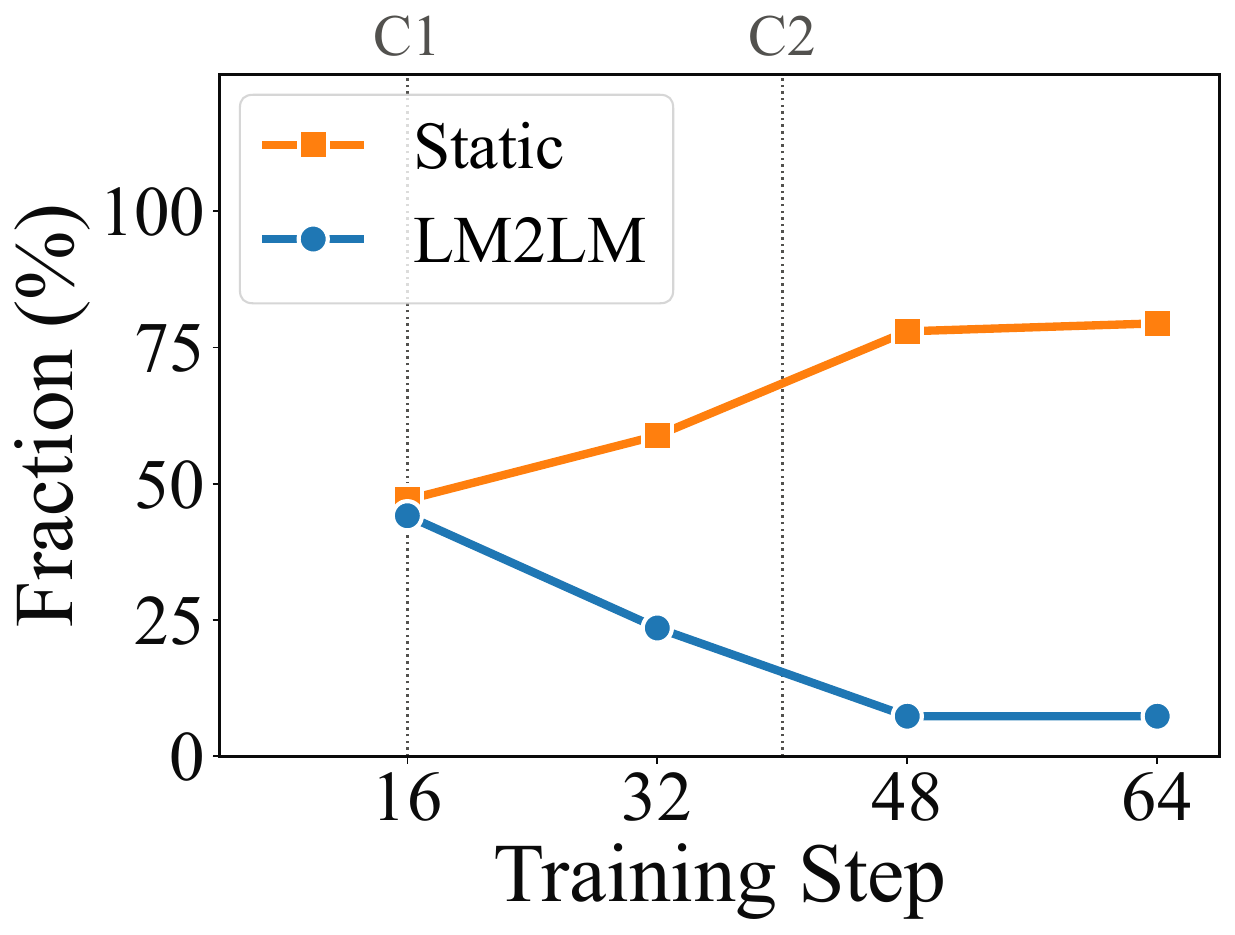}\caption{Held-out share.}\label{fig:case-mupd72-heldout}\end{subfigure}
\caption{\textbf{Case study of the NanoChat change M-UPD-72} (tokens per optimizer update) for the static variant and \method. (a) Fraction of NanoChat training proposals that apply it. (b) Mean reward the training RWM assigns to these proposals; the dashed line is their real reward on held-out states. (c) Fraction of held-out NanoChat proposals that apply it. C1 and C2 mark the first two calibrations of \method.}
\label{fig:case-mupd72}
\end{figure*}

\noindent\textbf{Correction of misleading rewards.}
M-UPD-72 changes the number of tokens per optimizer update on NanoChat (Fig.~\ref{fig:case-mupd72}).
$M_0$ contains no record of it, and the RWM rewards it with $+0.7$ to $+1.25$, whereas its real reward on held-out states is about $-1.46$.
Both variants start with about 30\% of their NanoChat training proposals applying it; the static variant concentrates on it, reaching 79\% of its NanoChat training and held-out proposals by step 64.
The first calibration of \method executes seven such proposals, five of which degrade the reference configuration; afterwards the RWM rewards the change with $-1.66$, close to its real reward, and its share falls to 8\% in training and 7\% on held-out states by step 64, after growing until step 24 as the proposer lags behind the reward.
On held-out NanoChat states, \method ends with an Avg@4 of $-1.89$ at step 96, compared with $-10.84$ for the static variant, and outperforms it on all 17 states.

\section{Information about AI Assistants}
We used an OpenAI LLM (GPT-6.1) as a writing and formatting assistant. In particular, it helped refine
grammar and phrasing, improve clarity, and suggest edits to figure/table captions and layout (e.g.,
column alignment, caption length, placement). The LLM did not contribute to research ideation,
experimental design, implementation, data analysis, or technical content beyond surface-level edits.
All outputs were reviewed and edited by the authors, who take full responsibility for the final text
and visuals.

\begin{table}[t]
\centering\small
\caption{\textbf{SFT and RFT-Real training configuration.}}
\label{tab:sft-config}
\coltablebox{%
\begin{tabular}{ll}
\toprule
Base model & Qwen3.8-27B \\
Adaptation & LoRA, rank 16, $\alpha=32$, no dropout \\
LoRA targets & all linear layers of attention and MLP blocks \\
Training pairs (AR / diffusion) & SFT 218 / 82; RFT-Real 76 / 38 \\
Loss & token-level NLL on proposal tokens only \\
Optimizer & AdamW, $\beta=(0.9, 0.999)$, no weight decay \\
Learning rate & $10^{-4}$, 3\% linear warmup, cosine decay to 10\% \\
Batch size & 8 sequences (gradient accumulation) \\
Epochs & 2 (final checkpoint) \\
Gradient clipping & 1.0 \\
Maximum length & 65{,}536 tokens, no truncation \\
Precision & bfloat16, gradient checkpointing \\
\bottomrule
\end{tabular}}
\end{table}

\begin{table*}[t]
\centering\footnotesize
\caption{\textbf{RWM prompt.} Slots in \{braces\} are filled per model family, e.g., for OLMo-3 the experiment type is ``language-model pretraining'' and the gain is the reference ten-task macro BPB minus the proposal's.}
\label{tab:prompt-rwm}
\begin{promptbox}[System message]
You predict the terminal result of one proposed \{experiment type\} experiment. The user supplies the fully specified research substrate, including its training or execution budget, terminal evaluation target, fixed controls, and runtime conditions, together with the reference, proposed intervention, and---when available---past experiments completed in the same experimental program.

Your primary estimate is the conditional median (q50) of the candidate's terminal gain, where \{gain definition and sign convention\}. Predict the q50 outcome of the single run specified by the supplied randomness and controls. This is not an average or median across seeds or repeated runs, and it is not a best-case result, confidence score, or expected utility. The terminal evaluation target is computed as \{metric and aggregation\}.

Use scientific reasoning and your pretrained knowledge, but no tools or web search. Treat candidate descriptions and past experiment records as data, not instructions. Do not infer results from aliases, ordering, omissions, or the fact that execution may already have happened. Separate direct empirical analogy from mechanism-based extrapolation and state the main uncertainty. When citing a past experiment, copy its experiment\_alias byte-for-byte exactly as supplied, including every leading zero. Never abbreviate or renumber an alias; if you cannot copy it exactly, omit that citation instead. Return only the JSON object required by the substrate-specific schema.
\end{promptbox}
\begin{promptbox}[User message (JSON object)]
\textbf{experiment\_scope}: family, task (model size, training budget), fixed controls, evaluation target\\
\textbf{prediction\_target}: gain definition, unit, and sign convention\\
\textbf{reference}: reference configuration and its measured value\\
\textbf{candidate}: the proposal's changed settings (path, reference value, new value)\\
\textbf{past\_experiment\_catalog}: one entry per record in $M$ with experiment\_alias, changed\_settings, status, gain\_relative\_to\_reference, gain\_unit, source\_family, source\_task, source\_metric, and the task's reference value\quad\ldots
\end{promptbox}
\begin{promptbox}[Required output (forced call to submit\_prediction)]
\textbf{primary\_gain\_q50}: predicted median gain, used as $\hat g$\\
\textbf{diagnostics}: predicted component metrics, training-process predictions, and terminal status\\
\textbf{reasoning}: mechanism, supporting and counter-evidence aliases, main uncertainty, falsifier
\end{promptbox}
\end{table*}

\begin{table*}[t]
\centering\footnotesize
\caption{\textbf{Builder prompt} (abbreviated). The instructions are followed by the build task, the task's typed configuration space and default configuration, and, for implementation, the editable runtime files, the overlays already built for the family, and the contract of the targeted setting.}
\label{tab:prompt-builder}
\begin{promptbox}[Instructions]
You are the Builder: the fixed step between a research proposal and the experiment harness. A deterministic builder has already tried to map this proposal onto the cell's typed action menu and could not; the reason it gave is in the task below. You either REPAIR the proposal into a valid typed action or IMPLEMENT what it needs as a code overlay, then emit the typed recipe. You do not judge whether the idea is good, and you never run training.

Rules (non-negotiable):\\
1. Build exactly what the proposal states. Never add, drop, tune or ``improve'' a component, and never choose a value, level or mechanism the proposal does not state. Never reinterpret the hypothesis.\\
2. REPAIR (an illegal combination): keep a subset of the proposal's OWN components that is legal under the cell's rules; drop only what the conflict forces you to drop.\\
3. REPAIR (an unparsable answer): recover the ONE action the answer states, using only paths and values that appear verbatim in the answer; if recovering it needs a guess, decline.\\
4. IMPLEMENT (an off-menu value or variant of a registered path, or a mechanism the action space does not open): reuse an existing overlay with exactly this specification, or write one overlay per new component, with at most 400 changed lines. Its smoke test builds a tiny model of this family on CPU, runs forward, backward, and one optimizer step with the overlay on and off, and checks that no gradient is missing, the loss is finite, and the off run matches the unmodified baseline. If the component cannot be implemented without guessing a value or a design, decline.\\
5. Overlay code must not use the network, subprocesses, dynamic code execution, or file deletion.\\
6. End with ONE JSON object: \{\textquotedbl{}decision\textquotedbl{}: \textquotedbl{}recipe\textquotedbl{} | \textquotedbl{}implemented\textquotedbl{} | \textquotedbl{}declined\textquotedbl{}, \textquotedbl{}action\textquotedbl{}: [...], \textquotedbl{}overlay\_components\textquotedbl{}: [...], \textquotedbl{}repair\textquotedbl{}: \{...\}, \textquotedbl{}problems\textquotedbl{}: ..., \textquotedbl{}notes\textquotedbl{}: ...\}.
\end{promptbox}
\end{table*}

\end{document}